\documentclass[12pt]{article}
\usepackage{framed}
\usepackage{algorithm}
\usepackage[margin=10in,bottom=10in]{geometry}
\usepackage[noend]{algpseudocode}
\usepackage[english]{babel}
\usepackage[utf8x]{inputenc}
\usepackage[T1]{fontenc}
\usepackage{apacite}
\usepackage[svgnames]{xcolor}
\usepackage{booktabs}
\usepackage{multirow}
\usepackage{siunitx}
\usepackage{algpseudocode}
\usepackage{mathrsfs}
\usepackage{listings}
\usepackage{color}
\usepackage{bbm}
\usepackage{adjustbox}
\usepackage{epsfig, epsf, graphicx}
\usepackage{tikz}
\usepackage{animate}
\usepackage{pstricks, pst-node, psfrag}
\usepackage{amssymb,amsmath,bm}
\usepackage{verbatim,enumerate}
\usepackage{rotating, lscape}
\usepackage{diagbox}
\usepackage[doublespacing]{setspace}
\usepackage{tikz}

\usetikzlibrary{arrows,calc,tikzmark}
\usepackage[colorlinks=true,linkcolor=blue,citecolor=blue]{hyperref}%
\usepackage{float}
\usepackage[small]{caption}
\usepackage{subcaption}
\usepackage{natbib}
\usepackage{lipsum}
\usepackage{threeparttable}
\usepackage{multirow}
\usepackage{multicol}
\usepackage[toc,page]{appendix}

\makeatletter
\def\BState{\State\hskip-\ALG@thistlm}
\makeatother
\makeatletter
\newcounter{phase}[algorithm]
\newlength{\phaserulewidth}
\newcommand{\setphaserulewidth}{\setlength{\phaserulewidth}}

\makeatother
\setphaserulewidth{.7pt}

\newcounter{case}[algorithm]
\newlength{\caserulewidth}
\newcommand{\0}{\mathbf{0}}

\newcommand{\bA}{\mathbf{A}}

\newcommand{\bepsilon}{\bm{\epsilon}}

\newcommand{\bH}{\mathbf{H}}
\newcommand{\bI}{\mathbf{I}}

\newcommand{\bOmega}{\bm{\Omega}}

\newcommand{\bR}{\mathbf{R}}
\newcommand{\bSigma}{\bm{\Sigma}}

\newcommand{\btheta}{\bm{\theta}}

\newcommand{\bW}{\mathbf{W}}
\newcommand{\bx}{\mathbf{x}}

\newcommand{\bX}{{\mathbf{X}}}

\newcommand{\bY}{\mathbf{Y}}
\newcommand{\bz}{\mathbf{z}}
\newcommand{\bZ}{\mathbf{Z}}

\newcommand{\R}{\mathbb{R}}
\newcommand{\setcaserulewidth}{\setlength{\caserulewidth}}
\newcommand{\mb}{\mathbf}
\newcommand{\bs}{\mb{s}}

\newcommand{\bsy}{\boldsymbol}

\makeatother
\setcaserulewidth{.7pt}
\algdef{SE}[SUBALG]{Indent}{EndIndent}{}{\algorithmicend\ }%
\algtext*{Indent}
\algtext*{EndIndent}

\def\diag{\hbox{diag}}

\def\boxit#1{\vbox{\hrule\hbox{\vrule\kern6pt
			\vbox{\kern6pt#1\kern6pt}\kern6pt\vrule}\hrule}}

\def\bse{\begin{eqnarray*}}
	\def\ese{\end{eqnarray*}}
\def\be{\begin{eqnarray}}
\def\ee{\end{eqnarray}}
\def\bq{\begin{equation}}
\def\eq{\end{equation}}
\def\bse{\begin{eqnarray*}}
	\def\ese{\end{eqnarray*}}

\begin{document}

\thispagestyle{empty} \baselineskip=28pt \vskip 5mm
\begin{center} {\Huge{\bf Modeling High-Resolution Spatio-Temporal Wind with Deep Echo State Networks and Stochastic Partial Differential Equations}}
	
\end{center}

\baselineskip=12pt \vskip 10mm

\begin{center}\large
Kesen Wang\footnote[1]{\baselineskip=10pt Statistics Program,
King Abdullah University of Science and Technology,
Thuwal 23955-6900, Saudi Arabia.\\
E-mail: kesen.wang@kaust.edu.sa, marc.genton@kaust.edu.sa\\
This research was supported by the
King Abdullah University of Science and Technology (KAUST).}, Minwoo Kim\footnote[2]{\baselineskip=10pt Department of Statistics, Pusan National University, Busan, 46241, South Korea. E-mail: mwkim@psuan.ac.kr}, Stefano Castruccio\footnote[3]{\baselineskip=10pt Department of Applied and Computational Mathematics and Statistics, University of Notre Dame,
Notre Dame, IN, 46556, USA.
E-mail: scastruc@nd.edu} and Marc G.~Genton\textcolor{blue}{$^1$}
\end{center}
\baselineskip=17pt \vskip 10mm \centerline{\today} \vskip 15mm

\begin{center}
{\large{\bf Abstract}}
\end{center}
In the past decades, clean and renewable energy has gained increasing attention due to a global effort on carbon footprint reduction. In particular, Saudi Arabia is gradually shifting its energy portfolio from an exclusive use of oil to a reliance on renewable energy, and, in particular, wind. Modeling wind for assessing potential energy output in a country as large, geographically diverse and understudied as Saudi Arabia is a challenge which implies highly non-linear dynamic structures in both space and time. To address this, we propose a spatio-temporal model whose spatial information is first reduced via an energy distance-based approach and then its dynamical behavior is informed by a sparse and stochastic recurrent neural network (Echo State Network). Finally, the full spatial data is reconstructed by means of a non-stationary stochastic partial differential equation-based approach. Our model can capture the fine scale wind structure and produce more accurate forecasts of both wind speed and energy in lead times of interest for energy grid management and save annually as much as one million dollar against the closest competitive model.

\baselineskip=14pt

\par\vfill\noindent
{\bf Some key words:} Echo State Network; Spatio-Temporal Model; Stochastic Partial Differential Equation; Support Points;  Wind Energy

\clearpage\pagebreak\newpage \pagenumbering{arabic}
\baselineskip=26pt

\section{Introduction}
The constantly hiking carbon footprint in the past decade has caught global attention. Emerging countries such as Saudi Arabia, whose economy has been so far almost exclusively focused on non-renewable energies, announced `Vision 2030', a comprehensive blueprint outlining measures to transition into an economy more focused on renewable energy. Saudi Arabia's economic development in the past century has been almost exclusively driven by its large oil reserves, which comprise of a quarter of the world's proven reserves and currently produce over a tenth of the annual world oil output \citep{nakov2013saudi}. To diversify the country's economy, recent efforts have been focused on reducing its excessive reliance on fossil fuels and invest in renewable energy. Indeed, `Vision 2030' has explicitly laid out a plan to construct wind turbines and contribute 16 GW to the domestic energy demand. Since at present there is little to no infrastructure for wind power in Saudi Arabia, understanding the local spatio-temporal structure of wind in order to locate and then forecast the most suitable construction sites is a strategic priority. { In wind energy markets for developed countries such as the United States, numerical weather models such as the High Resolution Rapid Refresh model (HRRR, \cite{dow22}) provides a continuous update of the wind (among other variables). In emerging countries such as Saudi Arabia, there is no cyberinfrastructure for a systematic and continuous update of the wind speed at high spatio-temporal resolution. Additionally, it is not possible to rely on observational data, which are too sparse and coarse to provide a country-level mapping of wind \citep{giani2020closing,crippa2021temporal}.} To address this, \cite{giani2020closing} provided {the first \textit{ad hoc} high resolution simulation performed over the country as no other existing dataset could provide the same level of information}, and used it to lay out a comprehensive plan to locate sites for wind turbine accounting for energy potential as well as associated maintenance costs. Additional work has focused on these sites to provide sensitivity study of the potential energy output against varying climate conditions \citep{zhang2021assessing} or disruption events associated with extreme conditions causing turbine shut-offs or failure \citep{chen2021assessing}. The said studies provide a first assessment to support the country's plan for a sustainable wind energy portfolio, however contingent political, economic and societal circumstances may be such that the optimal sites may not be eventually chosen for wind farming. As such, given the current lack of a definitive choice of sites for wind farming, a reliable forecasting approach to wind field across the entire country is still necessary.

{ Numerically simulated wind data is routinely used as input to statistical and machine learning models to obtain probabilistic forecasts for wind energy, see, e.g., \cite{xie22} for a comprehensive review. Relying entirely on numerical simulations would be computationally infeasible, as it would require an ensemble of simulations in real time for uncertainty quantification. As such, we aim at building a high-resolution spatio-temporal wind speed statistical model for prediction for the entire country which acts as a surrogate, as it is the practice in a vast body of work for other countries  \citep{esp22,sob24}.} Building such a statistical model is a very challenging task, given the considerable extension, as well as geographical diversity, of Saudi Arabia. Indeed wind exhibits nonlinear dynamics that traditional statistical models, such as autoregressive integrated moving average (ARIMA), but also generalized additive models \citep{hastie2017generalized}, generalized multiplicative models \citep{atto2016wavelet}, and generalized autoregressive conditional heteroskedasticity (GARCH, \cite{bollerslev1986generalized}) cannot capture efficiently.

Dynamical models relying on neural networks (henceforth referred to as machine learning models) are increasingly applied in modeling time series as alternatives to conventional statistical models. Indeed, neural networks can be arranged in a recursive structure (Recurrent Neural Networks, RNNs) which naturally adapts to time series data and have been shown be able to capture nonlinear dynamics competitively or even better than traditional statistical models \citep{bon24b}. RNNs are flexible but their very large parameter space coupled with the recursive nature of the model imply an unstable evaluation of the gradient and hence a challenging implementation of gradient-based optimization \citep{doya1992bifurcations}. To address this issue, \cite{jaeger2001echo} proposed an Echo State Network (ESN), a RNN with a reduced parameter space implied by sparse and random matrices, a feature greatly reducing the computational complexity for inference. A similar model where the matrix weights are also stochastic and sparse was proposed in case the input is (or can be converted to) spike trains (liquid state machines, \cite{maass2002real,bon24c}). 

In the case of a relatively small number of locations, ESNs could be merged with covariance-based geostatistical models \citep{bon23}, but larger data require some form of dimensional reduction in space. Recently, \cite{mcd17}, \cite{mcd19}, and  \cite{mcd19b} applied ESNs (both shallow and deep) under both frequentist and Bayesian frameworks to model large geostatistical data with uncertainty quantification. In order to reduce the spatial dimensionality, these works adopt principal component analysis, a linear transformation aimed at finding the main modes of variability. As an alternative approach \cite{huanghuang} proposed to select a collection of spatial knots, fit an ESN, forecast the values at the knots at then interpolate the remaining field with some covariance-based simple geostatistical technique. While representing an improvement, covariance-based interpolation (\textit{kriging}) would still suffer from scalability issues with very large spatial data, such as the one in our application, comprising of wind speed across a country the size of one fifth of the United States. 

In this work, we propose a model which achieves spatial dimension reduction for ESN forecasting in a more accurate and efficient manner, and hence provide more timely and accurate wind speed forecasting at the lead times of interest for wind energy applications. Our proposed method finds a set of representative knots by means of an energy distance-based approach. Then, instead of performing covariance-based interpolation, the knots are modeled via a Stochastic Partial Differential Equation (SPDE). In this work we rely on the equivalence of one of the most popular spatial models (the Mat\'en model, \cite{ste99}) to the solution from an advection-diffusion SPDE \citep{whi54}, and generalize the differential operator to allow for spatially varying coefficients which can account for a nonstationary behavior. The SPDE is discretized using a finite elements methods \citep{lindgren2011explicit}, which in this case is computationally convenient as the resulting model is a Gaussian Markov Random Field with sparse precision matrix. We also propose a new batch forecasting approach which is predicated on a sequential parameter re-estimation and allows for more accurate forecasting. Finally, inference is achieved using General Processing Unit (GPU) acceleration which will be shown to significantly improve the computational time. 

The paper is organized as follows. Section~\ref{sec:data} presents the wind speed simulations over Saudi Arabia. Section~\ref{sec:model} introduces the trend, temporal and spatial structure of the model. Section~\ref{sec:inference} details how to perform inference. In Section~\ref{sec:results} the model is applied to produce wind speed and energy forecasts. Section~\ref{sec:conclusion} concludes with a discussion.

\section{Wind Data} \label{sec:data}

In this work, we focus on hourly wind simulations from 2013 to 2016 in Saudi Arabia, using the Weather Research and Forecasting model (WRF; \cite{skamarock2008description}), with boundary conditions from the high-resolution operational European Centre for Medium-Range Weather Forecast Model. These simulations were conducted on a regular planar grid, and were originally introduced in \cite{giani2020closing} to assess the most suitable sites for wind farming. 

In total, four simulations were performed, with resolutions at 9km and 6km and two different planetary boundary layer schemes. The simulated data were validated against observations collected from the ten sites of King Abdullah City for Atomic and Renewable Energy monitoring network. In this work, we focus on the simulation with the Mellor–Yamada–Janji\'{c} planetary boundary layer scheme \citep{gomez2015sensitivity,janic2001nonsingular,mellor1982development} at 6km resolution, as this simulation was shown to be the closest to the actual observations \citep{giani2020closing}. In addition, we only consider the wind simulations on the mainland, excluding the offshore islands, for a total of $n=53,333$ locations. Every hour between January 1st 2013 to December 31th 2016 is considered, for a total of $T=35,040$ time points. All together, this represents nearly 2 billions space-time locations.

\begin{figure}[t!]
\centering
\begin{subfigure}{0.45\textwidth}
  \centering
  \includegraphics[width=0.87\textwidth,]{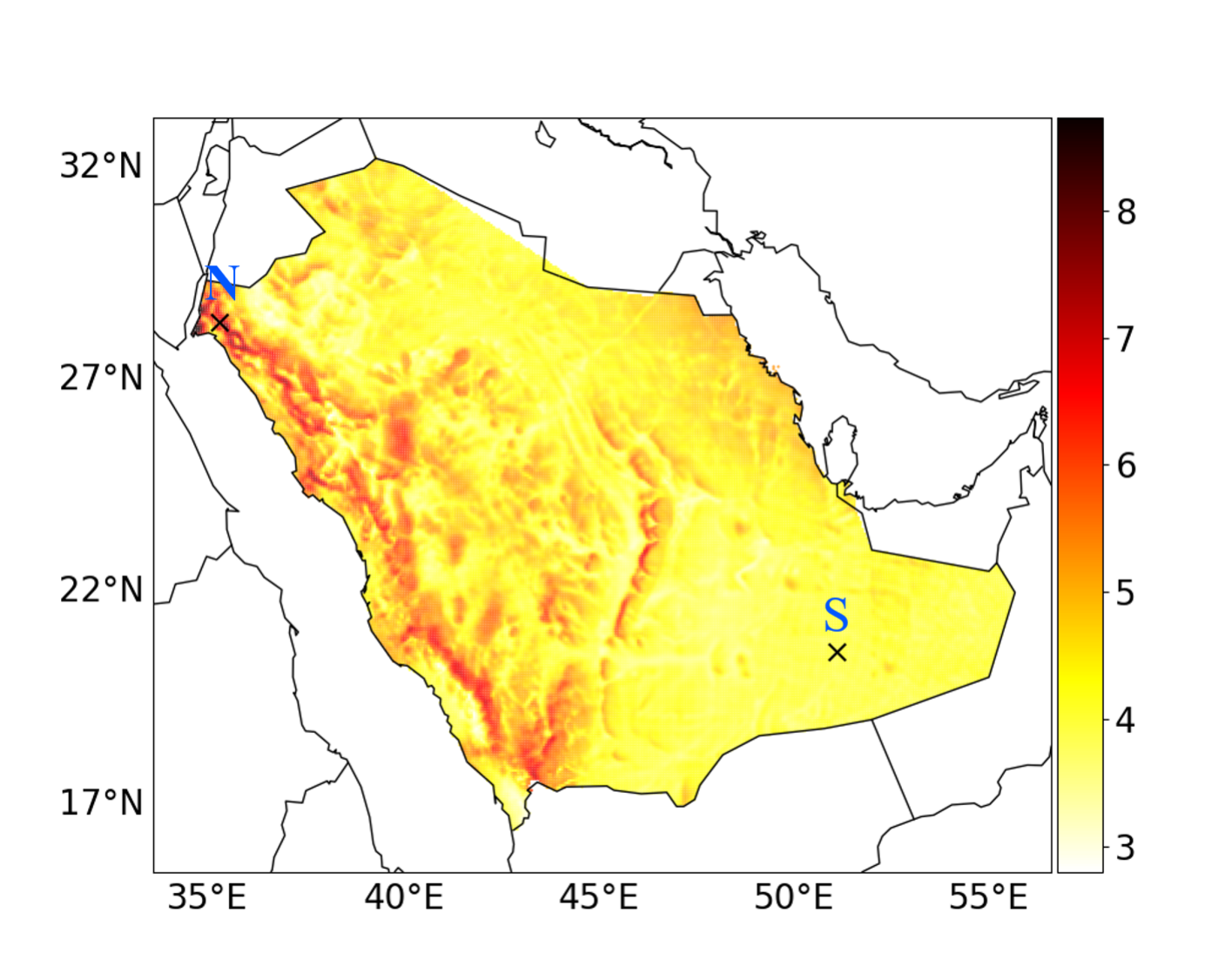}
  \caption{Mean}
  \label{F1_wind_mean}
\end{subfigure}
\begin{subfigure}{0.45\textwidth}
  \centering
  \includegraphics[width=0.87\textwidth,]{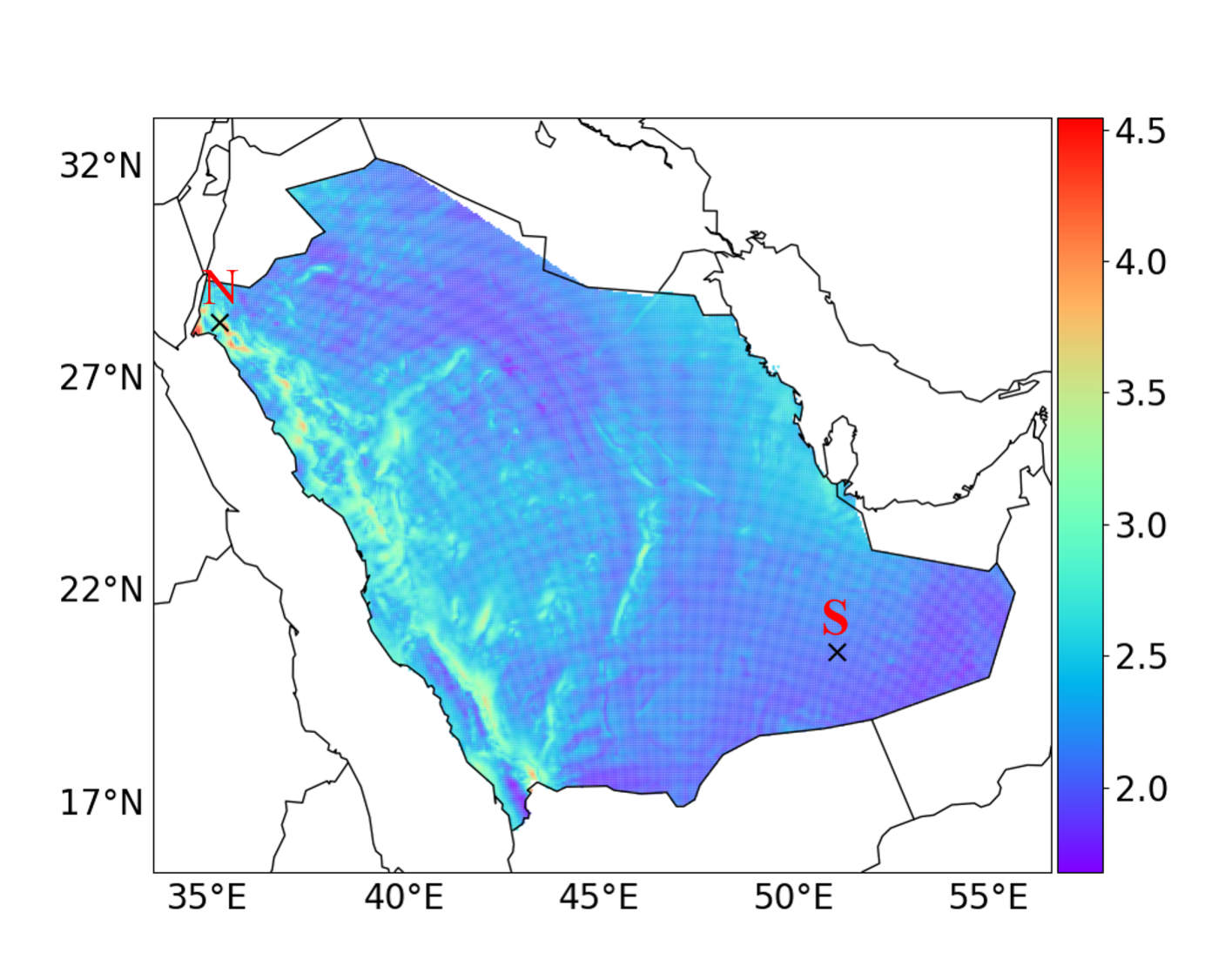}
  \caption{Standard Deviation}
  \label{F1_wind_sd}
\end{subfigure}
 \caption{(a) Mean and (b) standard deviation of the WRF simulated hourly wind speed (m/s) over Saudi Arabia from 2013 to 2016. The N and S are the two sample locations (marked in black crosses) used to compute the diagnostics in Figure \ref{FS2_acf}.}
 \label{fig:wind_data}
\end{figure}

The location-wise means and standard deviations of the wind speed data are shown in Figure~\ref{fig:wind_data}. Figure \ref{F1_wind_mean} indicates that the wind speed tends to be higher around the mountain ranges in the west of Saudi Arabia. Figure \ref{F1_wind_sd} showcases how the wind speed is more variable in the mountain ranges close to the west coast. Wind speed in the rest of the country is lower and more stable in comparison due to a flat topography, especially in the empty quarter (Rub' al Khali) in the South East. 

\section{The Spatio-Temporal Model} \label{sec:model}

We define $Z_t(\bs_i)$, where $i = 1,\dots,n$ and $t = 1, \dots, T$, as the wind speed observed at location $\bs_i$ and time $t$. Section \ref{sec:trend} provides a model for the trend. Section \ref{sec:spred} explains how the data in space are dimensionally-reduced. Sections \ref{sec:time} and \ref{sec:spatial} provide the temporal and spatial model, respectively.

\subsection{Trend Model}\label{sec:trend}

We model hourly wind speed for multiple years with a spatially varying harmonic trend:
\begin{equation}\label{eq:trend}
\sqrt{Z_t(\mb{s}_i)} = \beta_0(\mb{s}_i) + \sum_{p=1}^P \left\{\beta_{p,1}(\mb{s}_i)\text{cos}\left( \frac{2\pi t}{T_p} \right) + \beta_{p,2}(\mb{s}_i)\text{sin}\left( \frac{2\pi t}{T_p}\right) \right\} + \gamma(\mb{s}_i)Y_t(\mb{s}_i),
\end{equation}
where $P$ represents the number of harmonics. The parameter $\beta_0(\bs_i), i = 1, \dots, n,$ is the spatially varying intercept, while $\beta_{p,1}(\bs_i)$ and $\beta_{p,2}(\bs_i), p = 1,\dots,P,$ denote the spatially varying coefficients for each harmonic pair at period $T_p$, and $\gamma(\bs_i)$ is the location-wise normalizing parameter. The residual process $Y_t(\bs_i)$ is assumed to be independent in time and with unit variance and zero mean. Following the preliminary analysis in \cite{huanghuang}, we choose $P=5$ with harmonics at periods $1$ year, $0.5$ year, 1 day, 12 and 8 hours, so that $T_p = 4380, 2920, 1460, 8$, and $4$, respectively. The square root of $Z_t(\bs_i)$ allows to obtain data which more closely resemble normality, as demonstrated in Figure S3 in the Supplementary Material of \cite{huanghuang}. The following sections focus on modeling the spatio-temporal structure of the wind speed residuals,~$Y_t(\bs_i)$.


\subsection{Spatial Dimensional Reduction}\label{sec:spred}

In general, it is impractical to model the entire vector of spatial wind speed residuals $Y_t(\bs_i)$ for $n=53,333$ locations. For the sake of modeling and scalability, it is convenient to reduce the spatial dimension $n$ to $n_{\text{red}}\ll n$ (`red' stands for `reduced') and in this work we rely on support points \citep{mak2018support} to select the most representative locations. 

The support point method selects a set of locations that optimizes the energy distance \citep{mak2018support,szekely2004testing} between an empirical and a prescribed distribution. Formally, we assume two distribution functions $F$ and $G$ with finite means and $\textbf{S},\textbf{S}^\prime\overset{\mathrm{i.i.d.}}{\sim} G$ and $\bX,\bX^\prime \overset{\mathrm{i.i.d.}}{\sim} F$, and we consider  the energy distance:
\[
    E(F,G) \equiv 2\mathbb{E}\|\bX - \textbf{S}\|_2 - \mathbb{E}\|\textbf{S} - \textbf{S}^\prime\|_2 - \mathbb{E}\|\bX - \bX^\prime \|_2.
\]
Given that we have no prior information on the true distribution of the locations, we set $G=F_n$, where $F_n$ is the empirical distribution function of $F$ for $\{\mb{s}_i\}_{i=1}^n$, and the energy distance becomes
\begin{align}
    E(F,F_n) = \frac{2}{n}\sum_{i=1}^n\mathbb{E}\|\bX-\bs_i\|_2 - \frac{1}{n^2}\sum_{i=1}^n\sum_{j=1}^n\|\bs_i - \bs_j\|_2 - \mathbb{E}\|\bX - \bX^\prime\|_2.
\end{align}
The support points are defined as the set of $n_{\text{red}}\ll n$ locations optimizing the energy distance:
\begin{align}
    \{\bs_i\}_{i \in \mathscr{S}_{\text{red}}} = \underset{\bx}{\text{argmin}} \text{ } E(F_{n_{\text{red}}},F_n), \label{optimize}
\end{align}
where $\mathscr{S}_{\text{red}}$ is the set of indices for the selected locations ($|\mathscr{S}_{\text{red}}|=n_{\text{red}}$). The energy distance is non-negative for any two different distribution functions and is zero when the two distribution functions are equivalent. Therefore, by minimizing the energy distance we find the set of points for which the empirical distribution $F_{n_{\text{red}}}$ can best represent the true distribution $F$. 

\cite{song2022large} theoretically proved that spatial interpolations based on support points converge to the true values at an optimal rate compared with those based on randomly or regularly spaced locations; the rationale behind the optimal convergence rate is that the locations selected with minimized energy distance to the empirical distribution of the locations can most efficiently represent the entire data, as will also be shown in the simulation studies in the Supplementary Material. In this work, we apply support point to our $n = 53,333$ locations, and we aim at finding the set of locations $\{\bs_i\}_{i \in \mathscr{S}_{\text{red}}}$, where $n_{\text{red}} = 3,173$ represents the number of selected knots (chosen to maintain consistency with the number of knots in \cite{huanghuang}) to be optimized in equation \eqref{optimize}, which can be achieved through sampling-based inference over the entire location set $\{\bs_i\}_{i=1}^n$ \citep{mak2018support}. To achieve spatial dimension reduction, we set $F=F_{n_\text{red}}$ in \eqref{optimize}. The set of selected knots $(\bs^\prime_1,\dots,\bs^\prime_{n_{\text{red}}})$ based on $\underset{\bx}{\text{argmin}}$ is not necessarily a subset of the original locations. Therefore, we assign locations in the original set that are the closest to $(\bs_1^\prime,\dots,\bs_{n_{\text{red}}}^\prime)$ and denote this set as  $\{\bs_i\}_{i \in \mathscr{S}_{\text{red}}}$, the support points; { see the Supplementary Material for the optimality study and analysis of the support-point algorithm. }
\vspace{-3mm}
\subsection{Temporal Model} \label{sec:time}

{ In this section, we revisit the deep ESN model with $d=1, \ldots, D$ layers introduced in \cite{jaeger2001echo} and then \citep{mcd17} to model the temporal dynamics of the wind speed residuals $\mb{Y}_t=(Y_t(\mb{s}_1), \ldots, Y_t(\mb{s}_{n_{\text{red}}}))^\top$. In the following sections we propose a novel approach which uses ESN in conjunction with a spatial SPDE model, which allows for capturing more complex spatial dependence structure while still allowing for scalable and computationally affordable inference. The deep ESN is defined as}:
\begin{subequations}\label{eqn:ESN}
\begin{align}
    \text{output:} \quad & \bY_t =  \sum_{d=1}^{D}\textbf{B}_d k({\textbf{h}}_{t,d}) 
     + \sum_{d=1}^{D}\textbf{B}_d^2 k({\textbf{h}}_{t,d} \odot {\textbf{h}}_{t,d})+\bepsilon_t, \label{ESN:out}\\ 
    \text{hidden state d:} \quad &\textbf{h}_{t,d} = (1 - \alpha)\mb{h}_{t-1,d} + \alpha {\omega}_{t,d}, \label{ESN:lay1}\\
    &{\omega}_{t,d} = f\left(\frac{\nu_d}{|\lambda_{W_d}|}\bW_d\textbf{h}_{t-1,d}+ \bW^{\text{in}}_d \Tilde{\textbf{h}}_{t,d-1} \right ),\quad \text{for $d>1$}, \label{ESN:hidden}\\
    \text{reduction $d-1$:} \quad & \Tilde{\textbf{h}}_{t,d} = Q(\textbf{h}_{t,d}), \quad \text{for $d>1$}, \label{ESN:activation} \\
    \text{input:} \quad & {\omega}_{t,1}=f\left(\frac{\nu_1}{|\lambda_{W_1}|}\bW_1\textbf{h}_{t-1,1}+ \bW^{\text{in}}_1 \bx_{t} \right ), \label{ESN:input} \\
    \text{matrix distribution:} \quad & W_{d_{i,j}} = \gamma_{i,j}^{W_d}g(\eta_{W_d}) + (1 - \gamma_{i,j}^{W_d})\delta_0, \label{ESN:sparse1}\\
    & W_{d_{i,j}}^{\text{in}} = \gamma_{i,j}^{W_d^{\text{in}}}g(\eta_{W_d^{\text{in}}}) + (1 - \gamma_{i,j}^{W_d^{\text{in}}})\delta_0, \label{ESN:sparse2}\\
    & \gamma_{i,j}^{W_d} \sim Bern(\pi_{W_d}), \quad \gamma_{i,j}^{W_d^{\text{in}}} \sim Bern(\pi_{W_d^{\text{in}}}). \label{ESN:sparse3}
\end{align}
\end{subequations}

This nonlinear dynamic state-space model is such that in equation \eqref{ESN:out} the $n_{\text{red}}$-dimensional vector $\bY_t$ is formulated as a linear combination of the state vectors $\textbf{h}_{t,d}$ and of their quadratic states at each layer $d= 1,\dots,D$.  A scale location standardization $k(\cdot)$ is applied so that there is the same range of variability across layers. The error vector $\bepsilon_t$ is assumed to follow a Gaussian distribution with some covariance structure. { The operator $\odot$ is the elementwise Hadamard product.} The state vectors $\textbf{h}_{t,d}$ are then expressed in equation \eqref{ESN:lay1} as a convex combination with parameter $\alpha$ (the \textit{leaking rate}) of the previous hidden state vector $\textbf{h}_{t-1,d}$ and a vector $\bsy{\omega}_{t,d}$ accounting for long-range dependence. 

In \eqref{ESN:hidden}, $\bsy{\omega}_{t,d}$ is described as a combination of the state at the previous layer $\textbf{h}_{t-1,d}$ weighted by a matrix $\textbf{W}_d$ {(transition weight matrix at the $d$-th layer)} and a dimensionally reduced state vector $\Tilde{\textbf{h}}_{t,d-1}$ from the previous layer weighted by another matrix $\textbf{W}^{\text{in}}_d$ {(input weight matrix at the $d$-th layer)}. The \textit{activation function} $f(\cdot)$ in \eqref{ESN:hidden} and \eqref{ESN:input} is an hyperbolic tangent function, but other options such as the rectified linear or the sigmoid function could be used \citep{goodfellow2016deep}. The matrices $\bW_d$ are then divided by their largest eigenvalue $\lambda_{W_d}$ to ensure the \textit{echo-state property}, i.e., an asymptotic loss of dependence from the initial conditions \citep{lukovsevivcius2012practical,jaeger2007echo}. The parameter $\nu_d$ allows to scale the aforementioned matrix.

The dimension reduction for $\Tilde{\textbf{h}}_{t,d-1}$ is specified in \eqref{ESN:activation}, through the function $Q(\cdot)$, extracting information from the state vector $\textbf{h}_{t,d}$ as inputs for the next layer, and in this work we use principal component analysis. In the case of the first layer $d=1$, equation \eqref{ESN:input} explains how $\omega_{t,1}$ depends on a vector of past observations $\bx_t=(\bY_{t-\tau},\dots,\bY_{t-m\tau})^\top$, where $\tau$ denotes the forecast lead time, and $m$ represents the number of input past observations \citep{mcd17}. 

The weight matrices $\bW_d$  in \eqref{ESN:hidden} and $\bW_{d}^{\text{in}}$ in \eqref{ESN:input} are both assumed to be random and sparse according to a spike-and-slab distribution as specified in equations \eqref{ESN:sparse1}, \eqref{ESN:sparse2} and \eqref{ESN:sparse3}. Indeed, in both $\textbf{W}_d$ and $\textbf{W}_d^{\text{in}}$, each entry { $\{W_{d}\}_{i,j}$} and { $\{W_{d}^{\text{in}}\}_{i,j}$} is the outcome of a Bernoulli distribution $\gamma_{i,j}^{W_d}$ and $\gamma_{i,j}^{W^{in}_d}$, respectively. If the entry is not zero, then it is drawn from $g$, a symmetric distribution about 0. The parameters $\eta_{W_d}$ and $\eta_{W_d^{\text{in}}}$ control the shape of the distribution, in this case they represent the width of a zero-centered uniform distribution. The parameters $\pi_{W_d}$ and $\pi_{W_d^{\text{in}}}$ control the matrix sparsity of $\textbf{W}_d$ and $\textbf{W}_d^{\text{in}}$ respectively. 

\begin{sloppypar}
In summary, except for the set of matrices $\mb{B}=\{\mb{B}_1, \ldots, \mb{B}_D\}$, the ESN has the following hyper-parameters: 
\[
\btheta=(\{n_{h,d},n_{\Tilde{h},d},\nu_d,\eta_{W_d},\eta_{W_d^{\text{in}}},\pi_{W_d},\pi_{W_d^{\text{in}}}, d=1,\ldots, D\},m,\alpha,D).
\]      
\end{sloppypar}

\subsection{Spatial Model} \label{sec:spatial}

In this section, we introduce the method used to model the spatial dependence structure of $Y_t(\mb{s}_i)$ at the $n_{\text{red}}$ knots. As we assume this process is independent and identically distributed in time, we drop the time index for simplicity and we assume the existence of a continuously indexed random field $Y(\bs)$. We assume that this field is the solution of the following reaction-diffusion SPDE \citep{lindgren2015bayesian}:
\begin{align}
   (\kappa^2(\mb{s}) - \Delta)^{\alpha/2} (\tau(\bs)Y(\mb{s})) = \bW(\mb{s}), \quad \bs \in \R^2, \quad \alpha = \nu + 1, \quad \kappa > 0, \quad \nu > 0,
    \label{eq:SPDE}
\end{align}
where $\Delta$ denotes the Laplacian operator, and $\bW(\mb{s})$ is a standard spatial Gaussian white noise process. If $\kappa^2(\mb{s})=\kappa^2$ and $\tau(\bs) = \tau$ are constant in space, the only stationary solution is a Gaussian random field with the Mat\'ern correlation \citep{whi54}, precision parameter $\tau$, scale parameter $\kappa$, and smoothness $\nu$ \citep{ste99}. The marginal variance can be obtained as $\frac{\Gamma(\nu)}{\Gamma(\alpha)4\pi\kappa^{2\nu}\tau^2}$, and if $\nu$ is an integer, the discrete solution of the SPDE \eqref{eq:SPDE} is a Gaussian Markov random field \citep{rue2005gaussian}, a structure which implies sparse precision matrices and hence fast likelihood evaluation and computationally convenient inference \citep{lindgren2011explicit}. 

To allow for non-stationary spatial processes, \cite{ingebrigtsen2014spatial} proposed to formulate precision and scale parameters as linear combinations of basis functions:
\begin{align*}
    \log(\tau(\mb{s})) & = b_0^{\tau}(\mb{s}) + \sum_{o=1}^p b_o^{\tau}(\mb{s})\theta_o, \\
    \log(\kappa(\mb{s})) & = b_0^{\kappa}(\mb{s}) + \sum_{o=1}^p b_o^{\kappa}(\mb{s}) \theta_o, 
\end{align*}
where $b_o^{\tau}(\cdot)$ and $b_o^{\kappa}(\cdot)$ are the Fourier basis functions with coefficient parameters $\theta_o$. 

We assume prior independence among the coefficient parameters $b_o^{\tau}(\cdot)$ and $b_o^{\kappa}(\cdot)$ by assuming independent and identically distributed Gaussian priors with mean 0 and unit variance. We have also tried penalized complexity priors \citep{sim17} with 0.7 probability of the corresponding standard deviation $(\sigma)$ being greater than 1 $(\mu)$ and we noticed minimal difference in the results, so we opted for the simpler choice. 

\section{Inference} \label{sec:inference}

\subsection{Trend Model}

For the trend model \eqref{eq:trend}, the coefficients $\beta_0(\mb{s}_i)$, $\beta_{o,1}(\mb{s}_i)$, $\beta_{o,2}(\mb{s}_i)$, and $\gamma(\mb{s}_i)$ are estimated by first assuming independence across space, and hence reducing the model for each location to a linear regression. As such, ordinary least squares can be performed  on the entire wind data. 

The hourly model residuals from 2013 to 2016 are then denoted as $\hat{Y}^{\text{trend}}_t(\mb{s}_i)$ and could be used to estimate the temporal and spatial dependence. Figure \ref{FS2_acf}(a-b) shows the auto-correlation function for $\hat{Y}^{\text{trend}}_t(\mb{s}_i)$ for the two selected locations marked with black crosses in Figure \ref{fig:wind_data}. It is readily apparent how the temporal dependence is strong and a model for the temporal structure is necessary.

\subsection{Temporal Model} \label{inf:time}

In order to perform inference on the ESN in equation \eqref{eqn:ESN}, we first assume that the collection of matrices $\mb{B}$ is known, and we perform cross-validation by using wind speed data from 2013 to 2015 as a training set and 2016 as a testing set. The hyper-parameter vector $\btheta$ is then chosen as the minimizer of the mean squared error on the testing data. In order to estimate $\mb{B}$ we perform penalized least squares via ridge regression to mitigate overfitting. In symbols, we denote $\bY=\bH\textbf{B} + \bepsilon$, with: \\
\begin{align*}
   \bY = \begin{bmatrix} \bY_1^\top \\
   \vdots\\
   \bY_{T}^\top
   \end{bmatrix}, \quad \bH = \begin{bmatrix} \textbf{h}_1 & , &  {\textbf{h}_1^2}\\
   \vdots &  & \vdots\\
   \textbf{h}_{T} & , & {\textbf{h}_{T}^2}
   \end{bmatrix}, \quad \bepsilon = \begin{bmatrix} {\bepsilon_1}^\top \\
   \vdots\\
   \bepsilon_{T}^\top
   \end{bmatrix},
\end{align*}
where $\bY_t=(\mb{Y}_t(\mb{s}_1),\ldots, \mb{Y}_t(\mb{s}_{n_{\text{red}}}))$. In addition, the state vector at time point $t$, $\textbf{h}_t = \{k(\textbf{h}_{t,D}),k(\Tilde{\textbf{h}}_{t,D-1}),\dots,k(\Tilde{\textbf{h}}_{t,1})\}^\top$ and ${\textbf{h}_t^2} = \{k(\textbf{h}_{t,D} \odot \textbf{h}_{t,D}),k(\Tilde{\textbf{h}}_{t,D-1} \odot \Tilde{\textbf{h}}_{t,D-1}),\dots,k(\Tilde{\textbf{h}}_{t,1} \odot \Tilde{\textbf{h}}_{t,1})\}^\top$, where $t=1,\dots,T$. Then, we estimate the collection of coefficient matrices $\hat{\textbf{B}}$ via ridge regression is $(\bH^\top\bH + \lambda\bI)^{-1}\bH^\top\bY$ for a penalty $\lambda$ estimated via cross-validation.   

To test the ability of our deep ESN model in capturing the temporal dynamics, we calculate the difference in $L^2$ norms of the auto-correlation function (up to 50 time lags) between $\hat{Y}^{\text{trend}}_t(\mb{s})$ and the residuals after the ESN predictions, which we denote as $\hat{Y}^{\text{trend+time}}_t(\mb{s})$. Figure \ref{FS2_acf}(c-d) shows this improvement at the two sample locations (indicated by the crosses in Figure \ref{fig:wind_data}) and Figure \ref{ACF_diff} maps the difference in $L^2$ norms between the model with and without the temporal structure. It is readily apparent how there is a clear reduction with the medians (IQRs) of $1.110 (0.307)$, $1.212 (0.279)$, and $1.074 (0.270)$ for the three lead hours, respectively. 

\subsection{Batch Forecasting} \label{batch-update}

\begin{sloppypar}
Once the collection of matrices $\mb{B}$ has been estimated, the lead-one forecast at time point $T+1$ at the selected knots can be formulated as $\hat{\bY}_{T+1}=\hat{\mb{B}}^\top\textbf{h}_{T+1}$. The long-lead forecasts at time $T+a$ can be obtained by iteratively using the predicted values $\hat{\bY}_{T+1}, \ldots, \hat{\bY}_{T+a-1}$ as part of the input vector $\bx_t$. Instead of estimating $\mb{B}$ once, we propose a sequential update after a given batch window $b$. In practice, for all $a$ of interest (in our application we have $a=3$) we compute the forecast for time $T+a$ for all points $T+1,\ldots, T+b$, and then $\mb{B}$ is recomputed for the following batch. As it will be shown in Sections \ref{sec:comp} and \ref{sec:results}, there is a tradeoff in the choice of the window size $b$ between computational efficiency and forecasting accuracy, as a large $b$ will result in fast but less accurate results. We henceforth denote this approach as \textit{batch ESN (B-ESN)}.    
\end{sloppypar}

\subsection{Computational Sensitivity} \label{sec:comp}

This section assessed the computational burden in the ESN model as a function of the number of hidden states $n_h$, number of locations $n_{\text{red}}$, the computing hardware (CPU, GPU) with the batch size $b$ fixed at the temporal length of the test year 2016 to avoid updating the weight matrix $\hat{\textbf{B}}$ during inference. As for the sensitivity analysis to the size of forecasting window, this section examines the computational complexity for varying values of $b$ on the $n_{\text{red}} = 3,173$ knots selected in Section \ref{sec:spred} under both CPU and GPU computing environments. For both experiments, we used a 40-core Intel Cascade Lake machine with four V100 GPUs for this assessment.
\begin{figure}[b!]
\centering
\begin{subfigure}{0.32\textwidth}
  \centering
  \includegraphics[width=1\textwidth,]{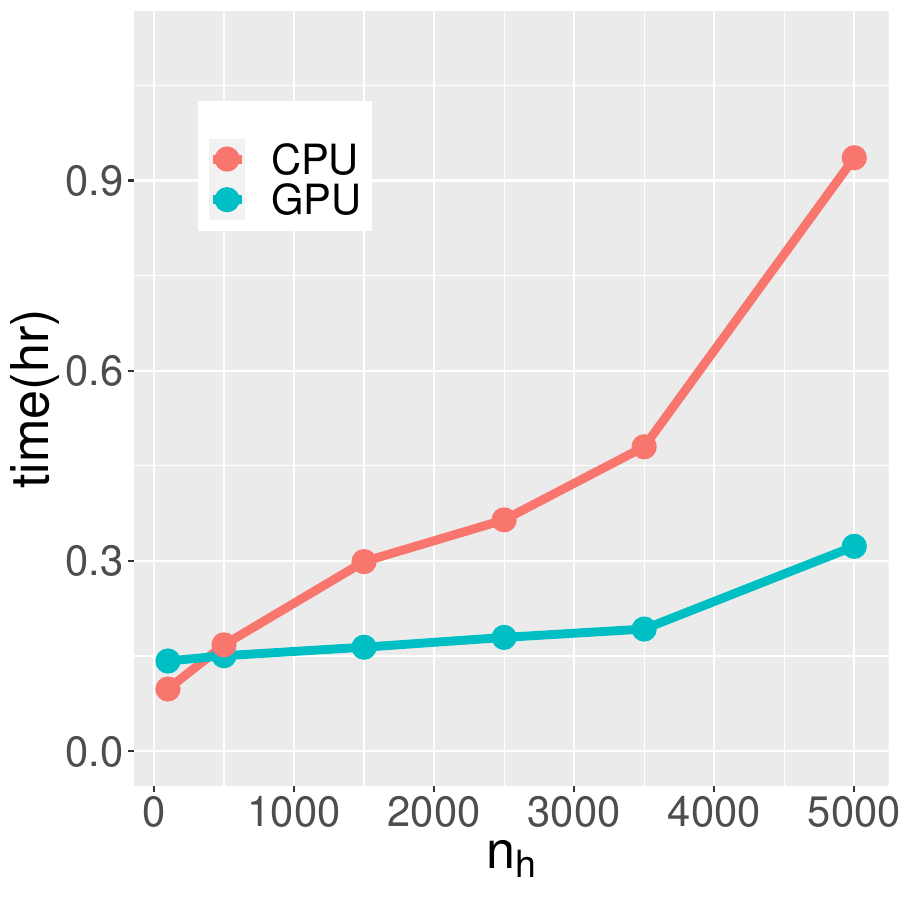}
  \caption{\hspace{5mm}$n_{\text{red}} = 100, b = 8,760$.}
  \label{F4_n=100_comp}
\end{subfigure}
\begin{subfigure}{0.32\textwidth}
  \centering
  \includegraphics[width=1\textwidth,]{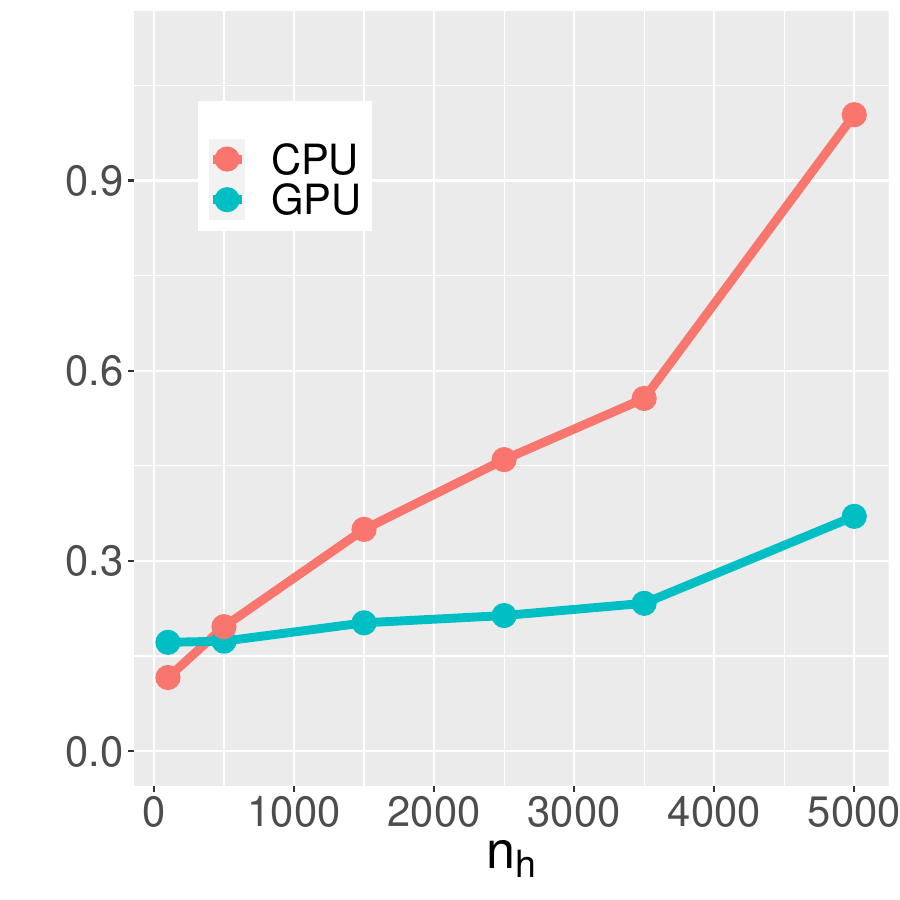}
  \caption{\hspace{5mm}$n_{\text{red}} = 3,173, b=8,760$.}
  \label{F4_n=3173_comp}
\end{subfigure}
\begin{subfigure}{0.32\textwidth}
    \centering
    \includegraphics[width = 1\textwidth,]{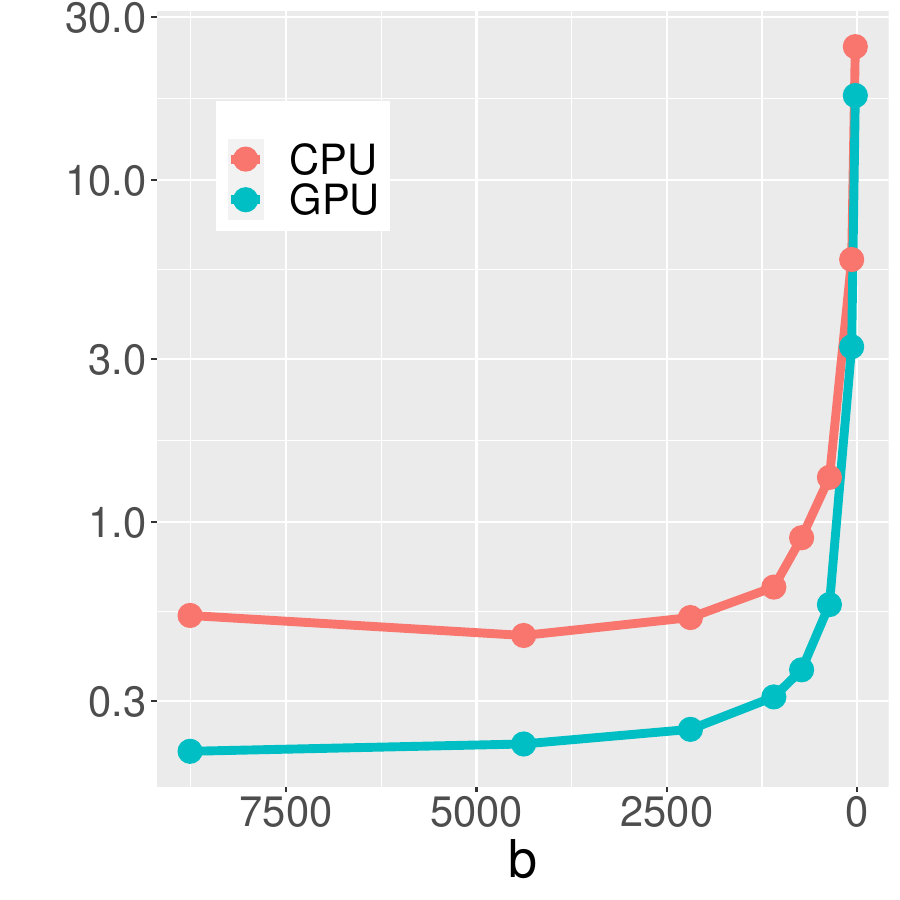}
    \caption{\hspace{6mm}$n_{\text{red}}=3,173.$}
    \label{F3_batch}
\end{subfigure}
\caption{Computation time for inference and forecasting { (three-hour ahead)} of the ESN model \eqref{eqn:ESN} as a function of the number of hidden state dimension $n_h$ for a small (panel (a)) and a large (b) number $n_{\text{red}}$ of knots. Panel (c) show the trade-off between prediction efficiency and computation complexity as a function of the batch size for forecasting; the order of the x-axis is reversed to show the increasing trend of computation time against the frequency of updates.}
\label{F4_comp}
\end{figure}
Figure \ref{F4_n=100_comp} and \ref{F4_n=3173_comp} indicate that the hidden state dimension ($n_h$) has more impact on the computation complexity than the number of knots ($n_{\text{red}}$) due to the use of quadratic dynamics \eqref{eq:SPDE}, for which the state vector $\textbf{h}_t$ contains both linear and second-degree states. Therefore, the corresponding transition weight matrix $\textbf{W}_d$ will scale as $O(4n_h^2)$. In contrast, as $n_{\text{red}}$ increases from 100 to 3,173, the computation time does not sensibly change for CPU and GPU computing time, which can be attributed to the $O(n_{\text{red}}^2)$ scaling rate of the associated input weight matrix $\textbf{W}_d^{\text{in}}$. Also, the CPU computing time rapidly increases as $n_{\text{red}}$ increases, while the GPU computing time is relatively constant. For large networks with $n_h = 5,000$, GPUs are considerably faster. If $n_{\text{red}}$ is very small, the GPU will require more computational time because of input/output latency outweighing any computational advantage. 

Figure \ref{F3_batch} shows the elapsed time for updating the weight matrix $\hat{\textbf{B}}$ with differing batch sizes. It is immediately apparent that weight updates are very computationally expensive as updating frequency increases. Both CPU and GPU consume a significant amount of time as the updating frequency rises. The sharp increasing trend is due to the constant re-assembling of the hidden state matrix $\bH$ and repeated re-estimating the coefficient matrix $\hat{\textbf{B}}$. Consequently, although more frequent updates of the coefficient matrix are beneficial for the prediction accuracy of long-range forecasts, as indicated by various slopes of the MSPE curves for the three lead hour forecasts shown in Figure \ref{F3_batch}, time complexity must be considered in actual practice to balance the trade-off between prediction accuracy and efficiency.   { In an operational setting, the inference time for $\textbf{B}$ cannot be comparable with the lead time for which the first forecast is sought. Figure~\ref{F3_batch} shows that when $b \approx 70$, the inference time for $\textbf{B}$ is around one hour. Therefore, in order for the hourly forecasts to be meaningful, we maintain $b > 70$ throughout the application.}

\subsection{Spatial Model} \label{inf:space}

In order to perform inference on the spatial model, the SPDE \eqref{eq:SPDE} must be solved. The domain is therefore divided into  a triangulated mesh containing non-intersecting triangles so that any two triangles on the mesh can have at most one overlapping corner or edge \citep{lindgren2011explicit}. We solve this SPDE by finite elements, assuming that the solution is of the form:
\begin{align}
    Y(\bs)=\sum_{o=1}^m\psi_o(\bs)Z_o, \label{basis}
\end{align}
where $\psi_o(\bs), o=1\dots,m$ are piecewise linear basis functions and $m$ is the number of vertices formulated within the mesh. In addition, $z_o$ are the associated weights following a Gaussian distribution. If $\nu$ is an integer or half integer, $\bZ=(Z_1,\dots, Z_m)^\top$ is a Gaussian Markov random field whose precision matrix $\bSigma_\bZ$ can then be approximated by the basis functions and their gradients. Moreover, $\bSigma_\bZ$ involves a local precision parameter $\tau(\bs)$ and a scale parameter $\kappa(\bs)$, controlling uncertainty and scales of spatial correlations. The SPDE model can be summarized as:
\begin{equation}\label{eq:spdeinf}
    \begin{array}{rcl}
         \bY|\bz,\sigma_e^2 &\sim & {\cal N}_{n}(\bA\bz,\sigma_e^2),  \\
         \bZ & \sim & {\cal N}_{m}(\0,\bSigma_{\bZ}).
    \end{array}
\end{equation}
In this model, $\bA$ is the projection matrix from the latent spatial process to the locations of interest, which enables predictions on the entire domain. Here, $\bZ$ represents the latent $m$-dimensional Gaussian Markov random field approximated by the piece-wise linear functions $\psi_o(\mb{s})$. The covariance $\bSigma_{\bZ}$ describes the dependence of the process $\bZ$, and since it is a Gaussian Markov random field, it allows fast likelihood evaluation with a sparse precision matrix. { The mesh grid is defined separately from the support points of the temporal model, whose observations are used for the estimation of the SPDE parameters. }

\subsection{Forecast Calibration} \label{sec:calibUQ}

Since the ESN assumes stochastic matrices, we can generate a forecast ensemble, with every member representing a different realization of the random matrices, whose uncertainty needs to be properly calibrated, so that a nominal 95\% prediction interval would contain approximately 95\% of the actual predictions. In this work, we adopt a calibration method as originally outlined in \cite{bon23}, based on adjusting the variances under the assumption of Gaussianity. We first group the predictions according to the prediction horizon $n_w$ and lead time $k$ as follows:
\begin{align*}
    \boldsymbol{f}_1^{(i)} = \begin{bmatrix}
        \hat{Y}_{T+1}^T(\mb{s}_i) \\
        \hat{Y}_{T+2}^{T+1}(\mb{s}_i)\\
        \vdots \\
        \hat{Y}_{T+n_w+1}^{T+n_w }(\mb{s}_i)
    \end{bmatrix}, \quad \boldsymbol{f}_2^{(i)} = \begin{bmatrix}
        \hat{Y}_{T+2}^T(\mb{s}_i) \\
        \hat{Y}_{T+3}^{T+1}(\mb{s}_i)\\
        \vdots \\
        \hat{Y}_{T+n_w +2}^{T+n_w }(\mb{s}_i)
    \end{bmatrix}, \dots, \boldsymbol{f}_k^{(i)} = \begin{bmatrix}
        \hat{Y}_{T+k}^T(\mb{s}_i) \\
        \hat{Y}_{T+k+1}^{T+1}(\mb{s}_i)\\
        \vdots \\
        \hat{Y}_{T+n_w+k}^{T+n_w }(\mb{s}_i)
    \end{bmatrix}, \label{eq16}
\end{align*}
where $\hat{\bY}_{T+k}^T=\mathbb{E}(\bY_{T+k}|\bY_T,\dots)$ and $i = 1,\dots,n$ indexes the location. In our application, we predict values three hours ahead for the entire year 2016 by setting $k=3$ and $n_w=8,757$ (the number of hours in the testing year 2016). Then the lead forecasts can be grouped as $\boldsymbol{f}_1^{(i)}=\left\{\hat{Y}_{T+1}^T(\mb{s}_i),\hat{Y}_{T+3}^{T+2}(\mb{s}_i),\dots,\hat{Y}_{T+n_w + 1}^{T+n_w}(\mb{s}_i)\right\}$, $\boldsymbol{f}_2^{(i)}=\left\{\hat{Y}_{T+2}^T(\mb{s}_i),\hat{Y}_{T+5}^{T+3}(\mb{s}_i),\dots,\hat{Y}_{T+n_w+2}^{T+n_w}(\mb{s}_i)\right\}$, and $\boldsymbol{f}_3^{(i)}=\left\{\hat{Y}_{T+3}^T(\mb{s}_i),\hat{Y}_{T+7}^{T+4}(\mb{s}_i),\dots,\hat{Y}_{T+n_w+3}^{T+n_w}(\mb{s}_i)\right\}$. The residuals are then computed group-wise using the observed values as: $\bR_j^{(i)}=\bY_j^{(i)} - \boldsymbol{f}_j^{(i)},j=1,2,3$, where $\bY_j^{(i)} = \{Y_{T+w}(\mb{s}_i);w=j,\dots,n_w+j\}$ and $\textbf{R}_j^{(i)} = \{R_j^{(i)}(w);w=j,\dots,n_w+j\}$. The empirical standard deviation of the residuals $\textbf{R}_j^{(i)}$ can be calculated as:
\begin{align*}
    \hat{\sigma}_j^{(i)} = \sqrt{\frac{1}{n_w}\sum_{w=j}^{n_w+j}\left(R_j^{(i)}(w) - \Bar{R}_j^{(i)}\right)^2}, \quad j=1,2,3,
\end{align*}
where $\Bar{R}_j^{(i)}=\sum_{w=j}^{n_w+j} R_j^{(i)}(w)/(n_w+1)$. If we formulate the calibrated residuals as 
$\Tilde{\textbf{R}}_{j,w} =(\Tilde{R}_j^{(1)}(w),\dots,\Tilde{R}_j^{(n)}(w))$, then 
these vectors are assumed to follow a multivariate normal distribution as a consequence of the Gaussianity of the ESN in Section \ref{inf:time}, i.e., $\Tilde{\textbf{R}}_{j,w} \overset{i.i.d.}{\sim} {\cal N}_{n}(\0,\bSigma)$ where independence is across $j=1,2,3$ and $w=j,\dots,n_w+j$. In order to estimate $\bSigma$, we use a generalized shrinkage method \citep{friedman2008sparse,castruccio2018scalable}, which assumes that:
\begin{equation}\label{eq:shrink}
\hat{\bSigma}^*(\delta) = \delta\hat{\bSigma}_{\text{SPDE}} + (1-\delta) \hat{\bSigma}_{\text{EMP}},    
\end{equation}
where $\hat{\bSigma}_{\text{SPDE}}=\bA\hat{\bSigma}_{Z}\bA^\top$ as defined from the inference on the SPDE model \eqref{eq:spdeinf}, and $\hat{\bSigma}_{\text{EMP}}$ is the empirical covariance matrix calculated from $\Tilde{\textbf{R}}_{j,w}$. The parameter $\delta \in (0,1)$ is chosen to make prediction intervals for the { spatial mean as close as possible to the expected coverage. This choice is justified by the fact that the variance of an average of dependent data is the most challenging quantity to estimate, as even a small misspecification of the covariance will propagate into this quantity}. The covariance matrix $\hat{\bSigma}^*(\delta)$ then characterizes the spatial correlation, which can be used to calculate the intervals for location-wise and aggregated forecasts. 

More general penalized quantile regression approaches as proposed in \cite{bon24} and then in  \cite{huanghuang} are possible, but only at the cost of an additional computational overhead which in this work was deemed excessive. 

\section{Application}\label{sec:results}

\subsection{Forecasting Results} \label{sec:forecastresults}

This section presents the forecasting results for our proposed model, considering the years 2013-2015 as the training set, and then 2016 as the testing set. The hourly wind predictions are performed on a rolling window basis, i.e., the forecast is continuously updated every hour. 

Table \ref{ESN_Forecast} shows the MSPE across the knots (first part) and across all locations (second part) for lead time from one to three hours across a selection of models similar to that of the simulation study in the Supplementary Material. In the first part of the table, only data at the knots are used, so no interpolation is necessary, while in the second part all locations are used (so the values are obtained through interpolation from the knots). As in the simulation studies, our model is denoted as B-ESN, while ESN denotes the one proposed in \cite{huanghuang}, in both the cases of $n_{\text{red}}$ knots and $n$ locations. PER denotes the persistence approach, while B-ESN-L denotes our model with interpolation performed with Lattice Kriging \citep{nyc15}. In the comparison study, we use two levels of multi-resolution basis to maintain the number of basis identical to the mesh vertices in the SPDE. Here GRU, LSTM, PER and VAR are the same models as in the simulation study in the Supplementary Material.
\begin{table}[t!]
    \centering
     \caption{Predictive comparison in terms of median MSPE { (computed across the temporal  dimension)} for different models (interquartile range or IQR in parenthesis). Results for one, two and three hours ahead forecasts at selected knots (first part of the table) and across all locations (second part) is shown. In bold is the name of our proposed approach and the minimum MSPE values across lead times. { Median and IQR in time are used as the temporal distribution of GRU, LSTM, PER, and VAR are skewed.} }
     \resizebox{.5\textwidth}{!}{
    \begin{tabular}{|c|c|c|c|}
    \hline
         Lead (hr) & One & Two & Three \\
         \hline
         \hline
         \multicolumn{4}{|c|}{Knots}\\
        \hline
        \textbf{B-ESN} &  0.065 (0.214) & \textbf{0.115} (0.355)& \textbf{0.157} (0.470)\\
        \hline
        ESN  & \textbf{0.064} (0.209) & 0.125 (0.386) & 0.173 (0.519)\\
        \hline
        GRU & 0.308 (0.888) & 0.380 (1.069) & 0.428 (1.194)\\
         \hline
        LSTM & 0.265 (0.743) & 0.351 (0.951) & 0.419 (1.127)\\
        \hline
        PER & 0.077 (0.277) & 0.198 (0.653) & 0.311 (0.976) \\
        \hline
        VAR & 0.087 (0.272) & 0.162 (0.486) & 0.216 (0.635)\\
        \hline
        \hline
        \multicolumn{4}{|c|}{All locations}\\
        \hline
        \textbf{B-ESN}  & 0.081 (0.267) & \textbf{0.133} (0.403) & \textbf{0.173} (0.510) \\
        \hline
        ESN  & 0.078 (0.255) & 0.141 (0.426) & 0.189 (0.557)\\
        \hline
        B-ESN-L & 0.088 (0.290)& 0.140 (0.425) & 0.181 (0.530) \\
        \hline
        PER & \textbf{0.077} (0.276 ) & 0.197 (0.669)& 0.310 (0.974) \\
        \hline
    \end{tabular}
    }
    \label{ESN_Forecast}
\end{table}

Compared to ESN, our B-ESN model relying on support points, batch update { $(b = 75)$} and SPDE for interpolation results in improved forecasts for two- and three-hour lead prediction horizon in terms of both accuracy and stability. Indeed, our model provides the lowest medians and IQRs for these two forecasting horizons both at the knots and at all locations. In addition, the one-hour lead forecasts obtained from B-ESN are very close to ESN for the knots and for all locations. 

\begin{figure}[t!]
\centering
\begin{subfigure}{0.35\textwidth}
\includegraphics[width=1\textwidth]{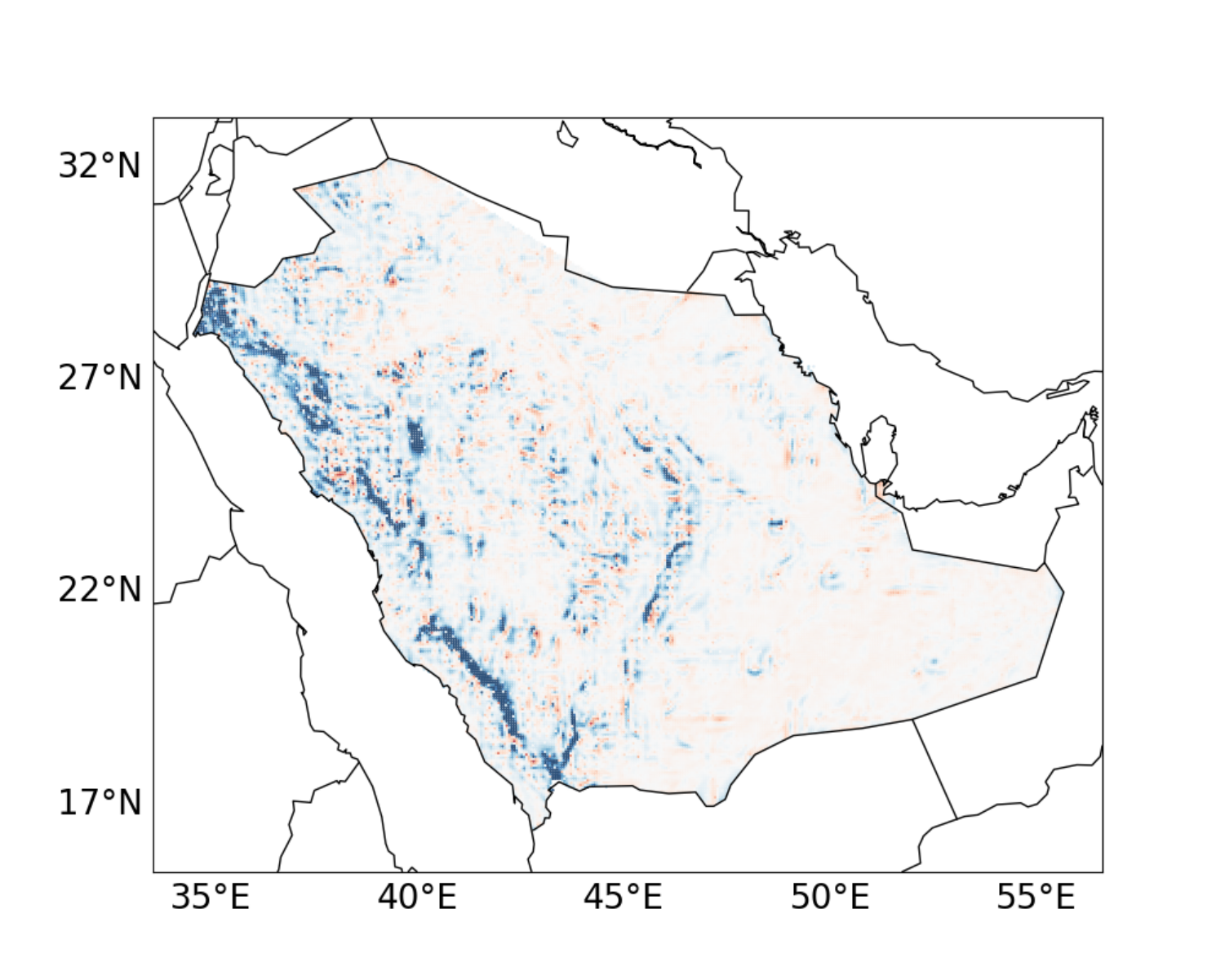}
\caption{One-hour lead}
\end{subfigure}
\hspace{-8.5mm}
\begin{subfigure}{0.35\textwidth}
\includegraphics[width=1\textwidth]{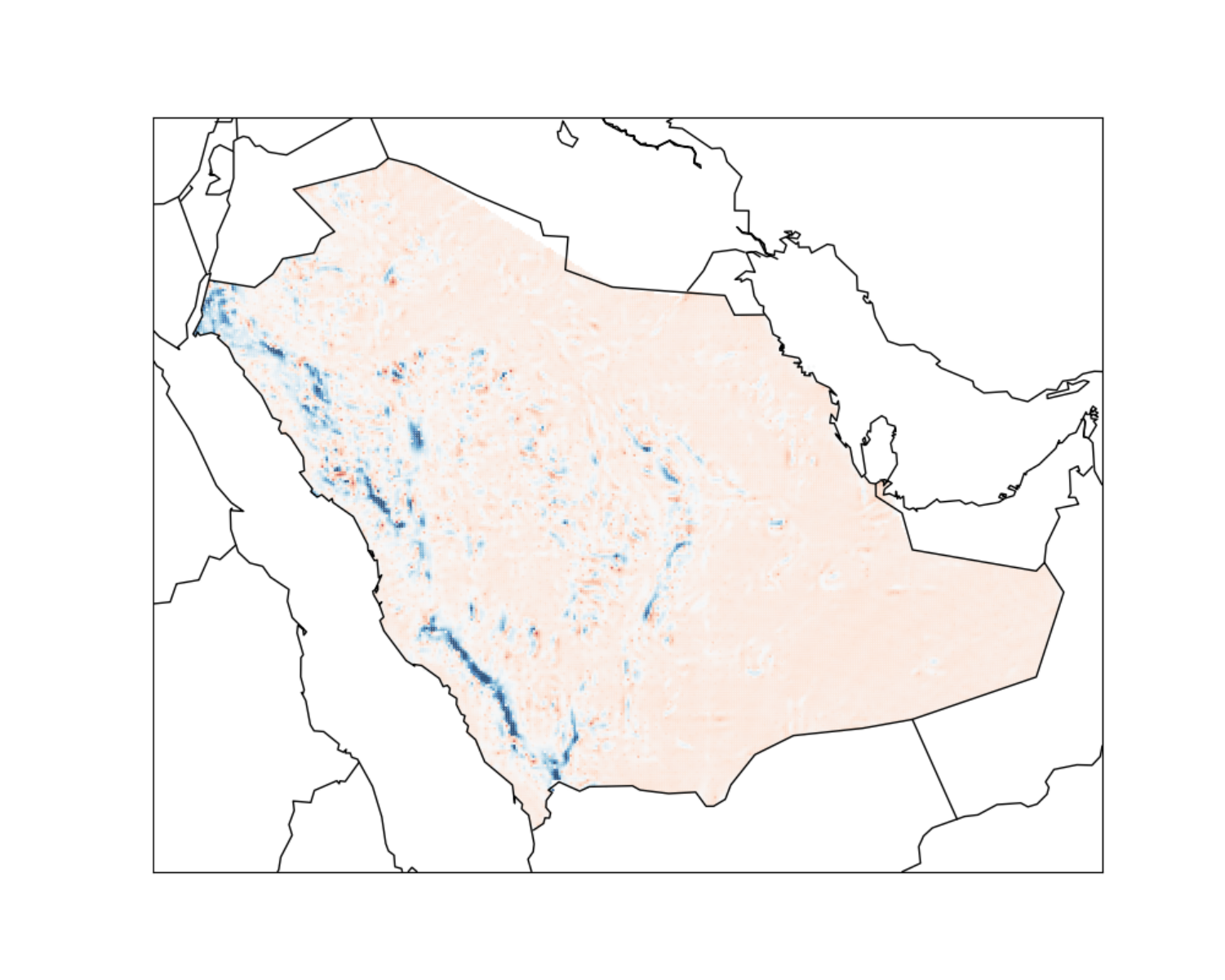}
\caption{Two-hour lead}
\end{subfigure}
\hspace{-8.5mm}
\begin{subfigure}{0.35\textwidth}
\includegraphics[width=1\textwidth]{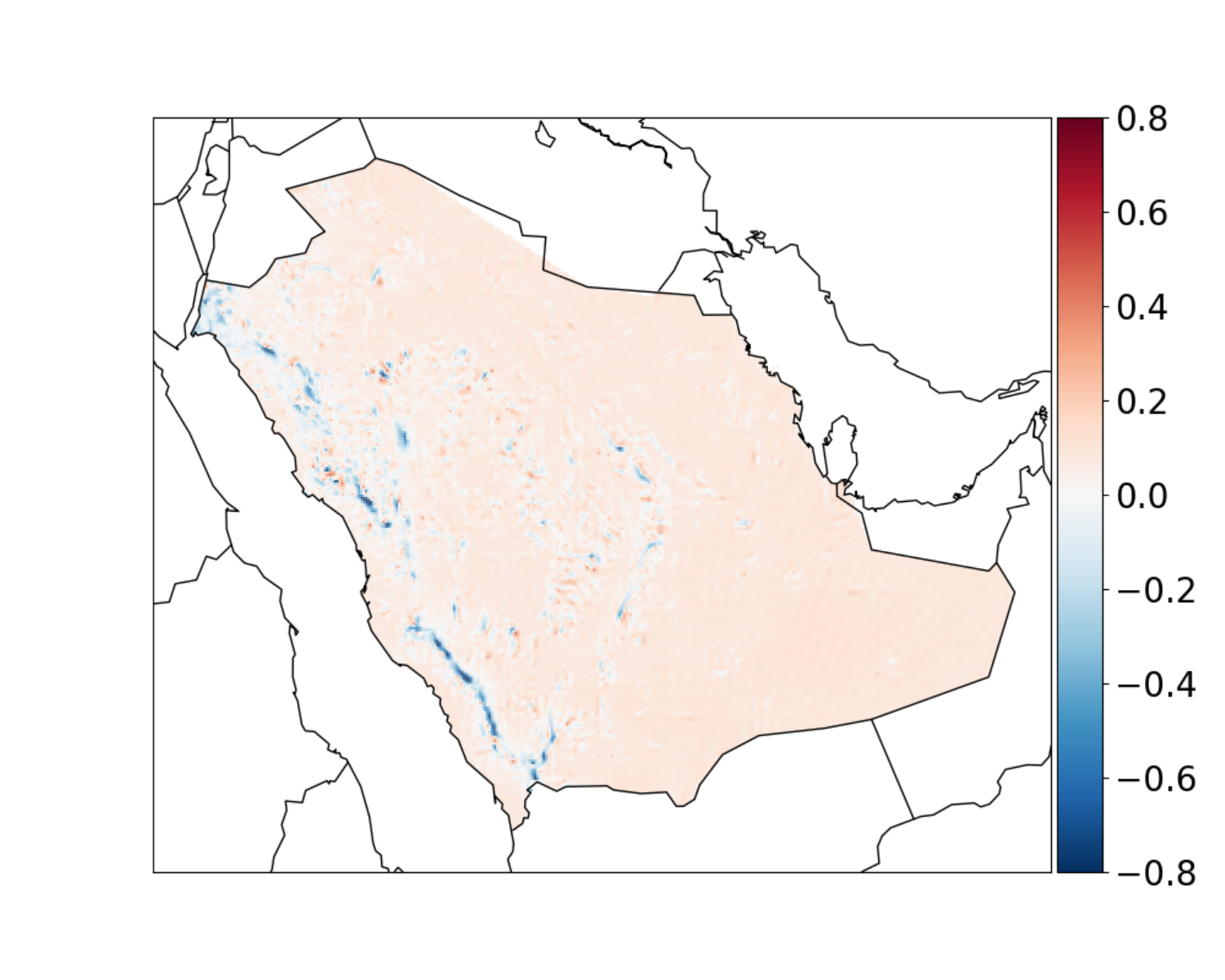}
\caption{Three-hour lead}
\end{subfigure}
\caption{Location-wise relative forecasting errors between ESN and B-ESN for up to three-hour lead ahead averaged across time. A positive number means that the B-ESN is better.}\label{fig:forclead}
\end{figure}

To better understand how B-ESN performs against ESN, Figure \ref{fig:forclead} maps the time-averaged relative forecasting error at all locations between these two models. Figure \ref{fig:forclead}a demonstrates that the forecasts from the two models are very similar overall for one-hour lead forecast, except in the Hejaz region (the western mountain ranges) where however wind farming is not practical due to logistics, construction costs and distance from to the country's energy grid \citep{giani2020closing}. Our B-ESN model outperforms the ESN for two and three { hours} ahead forecast, as shown in Figure \ref{fig:forclead}b and \ref{fig:forclead}c, in which the relative error becomes almost uniformly smaller.

In both the case of knots and all locations, Table \ref{ESN_Forecast} shows that our B-ESN model significantly outperforms the other statistical and machine learning methods chosen: PER, VAR, LSTM, and GRU (the MPSE is higher across all locations since the evaluation is over a larger number of points).

\subsection{Uncertainty Quantification} \label{sec:app_uq}

This section details the results from the calibration approach proposed in Section \ref{sec:calibUQ}. While marginally calibrating the forecast requires just a variance adjustment as indicated in Section~\ref{sec:calibUQ}, calibration of the joint distribution is a considerably more challenging task that in our model requires to choose the optimal convex combination between the covariance implied by the SPDE method and the empirical covariance, as shown in equation \eqref{eq:shrink}. { Since uncertainty quantification is more challenging for averages across locations (even a small covariance misspecification would be magnified for the average), focusing on such a large domain with $n=53,333$ locations is not feasible.}

Therefore, instead of focusing on uncertainty quantification for the grand mean for the entire domain, we focus on calibration for an expanding square. We choose a square which expands following a small increment horizontally and vertically from a reference point near the center of the domain, which in this case is 45$^\circ$ longitude and 20$^\circ$ latitude : $(45 - \Delta, 45 + \Delta) \times (20 - \Delta, 20 + \Delta)$, where $\Delta = 0.01,\dots,0.5$. We performed some preliminary tests of different initial locations and found comparable results. We then select the shrinkage parameter $\hat{\delta}(\Delta)$ that gives the best coverage for the grand mean of the forecasts within the expanding square, separately for each $\Delta$. Finally, among all choices of $\hat{\delta}(\Delta)$, we choose the one which gives the most accurate marginal coverage: the median (across all locations in the domain) empirical coverage of a 95\% needs to be as close as possible to the nominal 95\% value. For the ease of notation, we denote this estimated parameter as $\hat{\delta}$, so with the same hat notation but without the dependence on $\Delta$. Finally, this calibration is performed independently for each lead time, but again for the ease of notation we omit the dependence with respect to that. 

\begin{table}[t!]
    \centering
    \caption{Comparisons of 95\%, 80\%, and 60\% location-wise marginal coverage computed from the calibrated covariance $\hat{\bSigma}^*(\hat{\delta})$, $\hat{\bSigma}_{\text{spatial}}$ the covariance estimated from the SPDE model, and the sample covariance $\hat{\bSigma}_{\text{EMP}}$. The number here denotes the median of the empirical coverages across all locations and the number in parenthesis represents the IQR.}
    \resizebox{.5\textwidth}{!}{
    \begin{tabular}{|c|c|c|c|}
    \hline
    \multicolumn{4}{|c|}{Lead One} \\
     \hline
       EXPECTED & $\hat{\bSigma}^*(\hat{\delta})$  & $\hat{\bSigma}_{\text{spatial}}$ & $\hat{\bSigma}_{\text{EMP}}$ \\
         \hline
        0.95 & 0.949(0.022) & 0.765(0.049) & 0.977(0.013)   \\
         \hline
        0.80 & 0.813(0.041) & 0.569(0.042) & 0.876(0.037)\\
         \hline
        0.60 & 0.628(0.047) & 0.399(0.039)& 0.698(0.043) \\
         \hline
         \hline
          \multicolumn{4}{|c|}{Lead Two} \\
          \hline
           EXPECTED & $\hat{\bSigma}^*(\hat{\delta})$  & $\hat{\bSigma}_{\text{spatial}}$ & $\hat{\bSigma}_{\text{EMP}}$ \\
           \hline
           0.95 & 0.946(0.019) & 0.800(0.045) & 0.988(0.009) \\
           \hline
           0.80 & 0.809(0.039) & 0.606(0.049) & 0.906(0.030) \\
           \hline
           0.60 & 0.617(0.043) &0.431(0.040) & 0.738(0.044) \\
           \hline
           \hline
          \multicolumn{4}{|c|}{Lead Three}\\
          \hline
          EXPECTED & $\hat{\bSigma}^*(\hat{\delta})$  & $\hat{\bSigma}_{\text{spatial}}$ & $\hat{\bSigma}_{\text{EMP}}$\\
          \hline
           0.95 & 0.940(0.022)  & 0.834(0.042) & 0.992(0.005) \\
           \hline
           0.80 & 0.797(0.040) & 0.646(0.048) & 0.935(0.023) \\
           \hline
           0.60 & 0.604(0.047) & 0.457(0.046)& 0.780({ 0.045}) \\
           \hline
    \end{tabular}
    }
    \label{coverage}
\end{table}

Table \ref{coverage} shows how the calibrated covariance $\hat{\bSigma}^*(\hat{\delta})$ can achieve the expected coverage with stability accurately. It is readily apparent also how  $\hat{\bSigma}_{\text{spatial}}$ underestimates the coverage, while $\hat{\bSigma}_{\text{EMP}}$ significantly overestimates it. { This overestimation from $\hat{\bSigma}_{\text{EMP}}$ is due to the loss of local dependence structure assumed in the spatial model. Similarly, $\hat{\bSigma}_{\text{spatial}}$ underestimates the intervals because the assumption of the SPDE model on the entire study region, the entire country of Saudi Arabia in this case, is too strong. Hence, the dependence structure provided by $\hat{\bSigma}_{\text{spatial}}$ is not sufficient to account for variations on a global scale.  Such observations  are supported by the information provided in Table \ref{shrinkage}, demonstrating that $\hat{\delta}$ tends to decrease as the square expands (swinging from local to global dependence) and becomes very close to zero if the square is large enough.} The convex combination of the spatial and sample covariance allows to strike the balance between the regional spatial effects and the large-scale domain. It is also observable from Table \ref{shrinkage} that as forecasting lead increases, the convex combination tends to favor more the spatial covariance. This result is expected because lead forecasts are generated based on predicted values at the previous time point from the ESN model; therefore, we depend increasingly on the spatial model to interpolate the field instead of the original data. 

\subsection{Wind Power Prediction}\label{sec:windpow}

Electricity derived from wind power is challenging to store and must be consumed in a relatively short time. Therefore, ensuring accurate wind power forecasts is crucial for organizing the energy grid and estimate the required additional electricity from other energy sources in the near future. Wind power forecast is typically focused on short time horizons spanning from one to three hours, aiding in scheduling electricity transmission, allocating resources, and dispatching generated power \citep{gneiting2006calibrated,hering2010powering}. In this section, we convert our wind speed forecasts into wind power and quantify the advantages from an operational point of view. Saudi Arabia does not yet have a developed energy market with pricing for each lead time, so as reference we focus on the wind power predictions based on the two-hour lead forecasts. 

To compute the wind power, an extrapolation from the wind at the surface (i.e., at a height of 10 meters according to the WRF simulations) to the height of the turbine hub (80 to 120 meters depending on the model) must be performed. Many approaches are available in literature for vertical extrapolation of wind at hub height, see \cite{gua19} for a comprehensive review. The most popular approach assumes a power law relation at different heights as follows:
\begin{equation}\label{eq:extrap}
    Z_t^{(h)}(\bs) = Z_t(\bs)(h/10)^{\alpha(\bs)}\exp(\varepsilon_t),
\end{equation}
where $h$ denotes the hub height and $\alpha(\bs)$ is the \textit{wind shear coefficient}, a parameter controlling the rate of extrapolation and $\varepsilon_t\sim \mathcal{N}(0,\sigma^2)$ is the error independent and identically distributed in time and space. The simplest approach is to assume that the wind shear coefficient is constant in space and fixed at 1/7, under the assumption of near-neutral atmospheric stability on a flat terrain \citep{gua19, tagle2019non}. Recently, \cite{crippa2021temporal} proposed to estimate $\alpha(\bs)$, { validated the proposed model against winds at the surface and at 40, 60, 80 and 100 meters for the same data set that we are using here} and demonstrated significant improvement in the efficiency of wind power predictions. Therefore, here we use the same approach and fit the model in equation \eqref{eq:extrap} by performing a linear regression for every pixel independently on the logarithmic scale. Once extrapolated at hub height, the wind speed is converted into energy via a power curve. Each turbine model has a different shape of power curve, but share the same structure: a cut-in speed (minimum value for which power is produced), a ramp-up, and a maximum rated output power until a cut-out speed for which the turbine is shut down to prevent damage, see Figure \ref{fig:power_curve} for an example.

In this work, we focus on the 75 optimal locations for (inland) wind farming as indicated by \cite{giani2020closing}, which accounted for construction cost, maintenance, power lines construction and energy grid deployment; see Figure \ref{fig:power_loc}. For each of these locations an optimal turbine model was identified, and here we will use the corresponding power curves for these models.

\begin{table}[t!]
    \centering
      \caption{Annual sum of total absolute differences between the wind energy obtained using the forecasts from various models and true wind.}
    \begin{tabular}{|c|c|}
       \hline
       Model  & Absolute Energy Difference (KWh) \\
       \hline
       B-ESN  &  $2.601 \times 10^8$ \\
       \hline
       ESN & $2.770 \times 10^8$ \\
       \hline
       B-ESN-L & $2.995 \times 10^8$\\
       \hline
       PER & $3.120 \times 10^8$\\
       \hline
    \end{tabular}
    \label{wind_energy_diff_comp}
\end{table}
\begin{figure}[t!]
\centering
\begin{subfigure}{0.44\textwidth}
  \includegraphics[width=1\textwidth,]{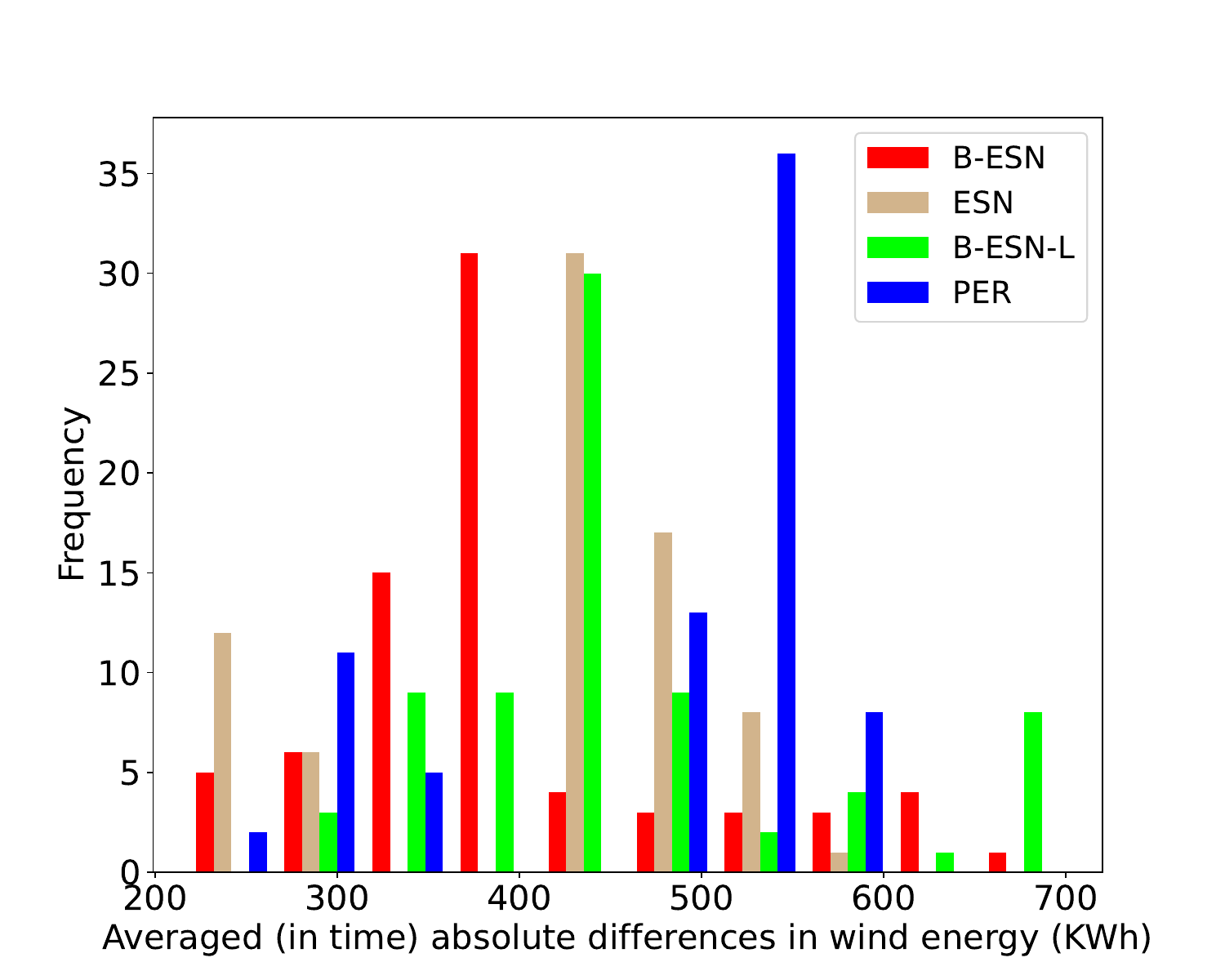}
  \caption{Model comparison}
  \label{F3_hist}
\end{subfigure}
\hspace{5mm}
\begin{subfigure}{0.44\textwidth}
  \includegraphics[width=0.93\textwidth,]{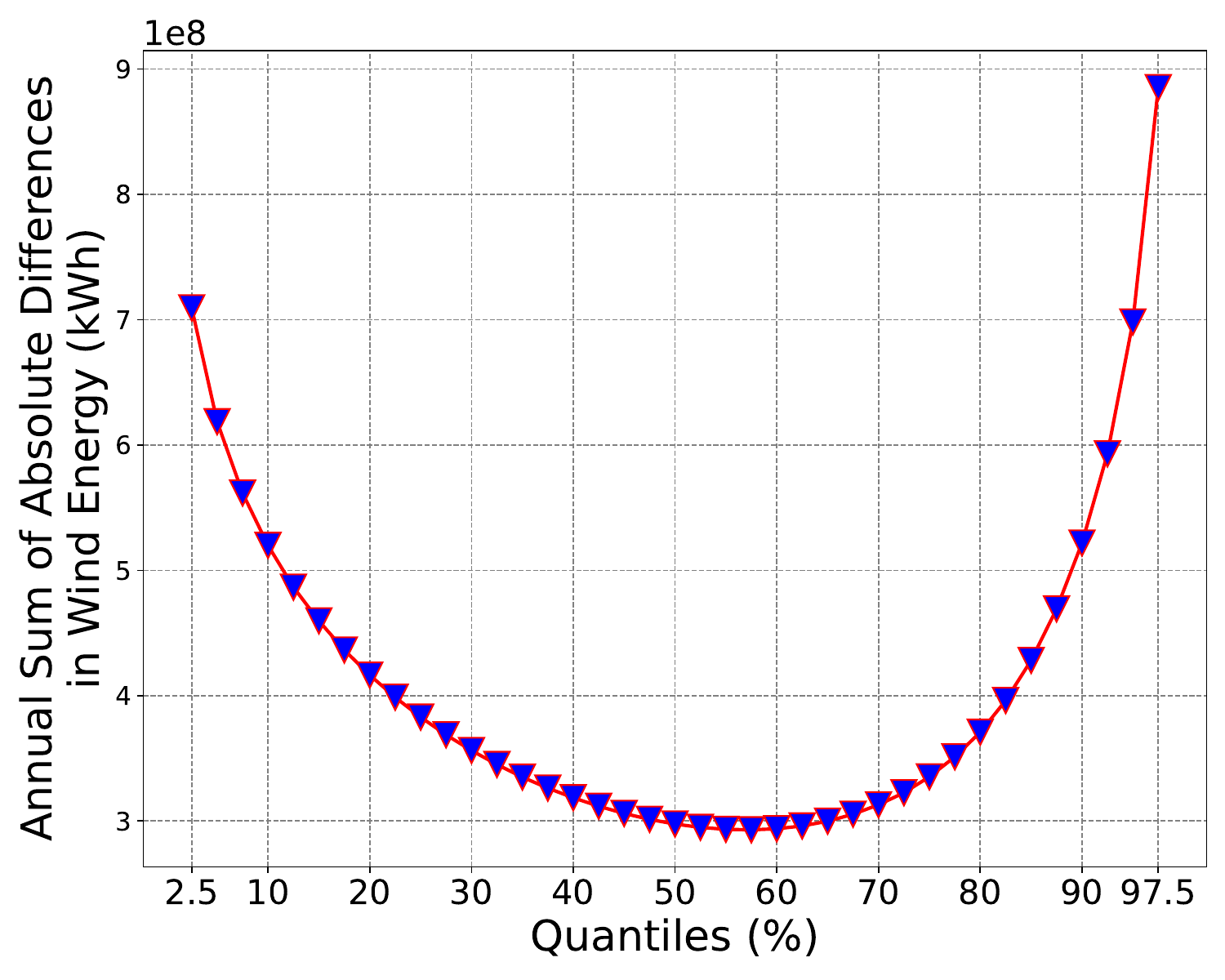}
  \caption{Quantile comparison}
  \vspace{-5mm}
  \label{F3_power}
\end{subfigure}
\caption{(a): Histograms of the absolute differences between the predicted wind power and the true wind power obtained from our B-ESN against ESN, B-ESN-L and PER. (b): Forecast quantiles against annual sum of the absolute differences between the B-ESN wind power predictions and the true wind power.}
\end{figure}

The annual sum of total absolute differences is computed and summarized in Table \ref{wind_energy_diff_comp}. It is readily apparent how the superior forecasting skills of B-ESN for wind speed against ESN, B-ESN-L and PER (the same models used in Table \ref{ESN_Forecast}) translate into superior forecasting skills for wind power. This is also evident from Figure \ref{F3_hist}, which shows the histogram of the absolute differences in wind energy of our B-ESN model against all aforementioned models.

Figure \ref{F3_power} demonstrates that the minimum annual sum of absolute differences is reached in the 50\% to 60\% interval and the difference remains stable within the 50\% to 70\% interval, implying that our combined model is resistant to overestimation of the wind speed. Moreover, as long as the forecast stays within the 20\% to 85\% interval, increases in the annual sum of absolute differences are limited by $1\times 10^8$ KWh, again testifying to the robustness of our model.  

Electricity costs in Saudi Arabia are currently estimated to be approximately 70.00 USD/MWh \citep{aldubyan2021energy}, considerably less expensive than the 173.00 USD/MWh in the United States (averaged across states, \cite{bls24}). Using these prices as reference, the proposed model could save as much as 1.183 million dollars per year compared to ESN, the second best model available, and more than 5 million dollars in stark contrast with the simplistic forecasts generated using persistence, { see \cite{hering2010powering} and \cite{zhu2012short} for detailed discussions regarding integrative challenges and various cost functions (symmetric and asymmetric) of transferring or storing the wind energy into the power system.}

\section{Conclusion}\label{sec:conclusion}

In this work, we propose a space-time model whose temporal part is controlled by an ESN, which is a sparse, stochastic RNN able to capture non-linear dynamics and long-range dependence for relatively short time series. Given the large size of the spatial data, dimension reduction is achieved via support points, which we showed in a simulation study to be able to better capture the spatial information than other sampling schemes. Finally, once the forecast is obtained, interpolation is performed using a non-stationary SPDE model, attaining overall higher prediction accuracy compared with previous approaches, as also shown in a simulation study. Overall, our proposed model allows for separate modeling of temporal and spatial dependence, and is more applicable to big datasets such as wind speed for a country as large as Saudi Arabia. 

When fitting the spatial model, we fixed the smoothness parameter $\nu = 0.5$ to ensure the Gaussian Markov property for computational efficiency. Although \cite{anynubolin} have implemented the SPDE approach for any arbitrary values of $\nu$ through covariance-based rational approximations of fractional SPDEs \citep{rspde}, applying this more general approach in this setting would likely lead to limited improvement in prediction accuracy because $\nu$ is estimated to be approximately 0.3 (close to 0.5) and the additional computational overhead of this approximation would hamper the efficiency of our operational forecasting approach, for which results must be provided in substantially less time than an hour. 

Given the importance of accurate forecast for wind power production, future work will investigate ensemble approaches to augment the forecast proposed here with other competitive approaches. In particular, ongoing work is focused on the use of 1) transformers and 2) convolutional auto-encoders as alternate means of spatial dimension reduction aimed at capturing more fine scale details of wind across Saudi Arabia. Additionally, better estimates of wind extrapolation to hub height (80 to 120 meters) could be provided if the WRF simulations were to have sufficiently close vertical levels. For numerical stability, however, this would be possible only with a much finer spatial and temporal resolution, with a substantially higher computational burden. Finally, estimates of monetary savings would require additional context and an assessment of the uncertainty, and to this end ongoing work is focused on characterizing and modeling the Saudi Arabia's energy market. 

\vspace{-.5cm}

\section*{Acknowledgements}
This publication is based upon work supported by King Abdullah University of Science and Technology Research Funding (KRF) under Award No. ORFS-2022-CRG11-5069.
\baselineskip=13pt

\bibliographystyle{apalike2}
\bibliography{references}

\begin{center}
\textbf{\large Data and Code Availability}
\end{center}
The data are available through the KAUST library at the following link: \url{https://repository.kaust.edu.sa/items/2187941f-4359-44b4-b593-1acd1e140d28}. The code for this work can be found at the following GitHub repository: \url{https://github.com/wangk0b/Deep-ESN-SPDE}.

\noindent

\newpage

\section*{Supplementary Material}
\setcounter{figure}{0}
\setcounter{table}{0}
\renewcommand{\thefigure}{S\arabic{figure}}
\renewcommand{\thetable}{S\arabic{table}}

\section*{Section 4: Inference}

\subsection*{Section 4.2: Temporal Model}
\begin{figure}[H]
\centering
\begin{subfigure}{0.45\textwidth}
  \centering
\includegraphics[width=0.8\textwidth,]{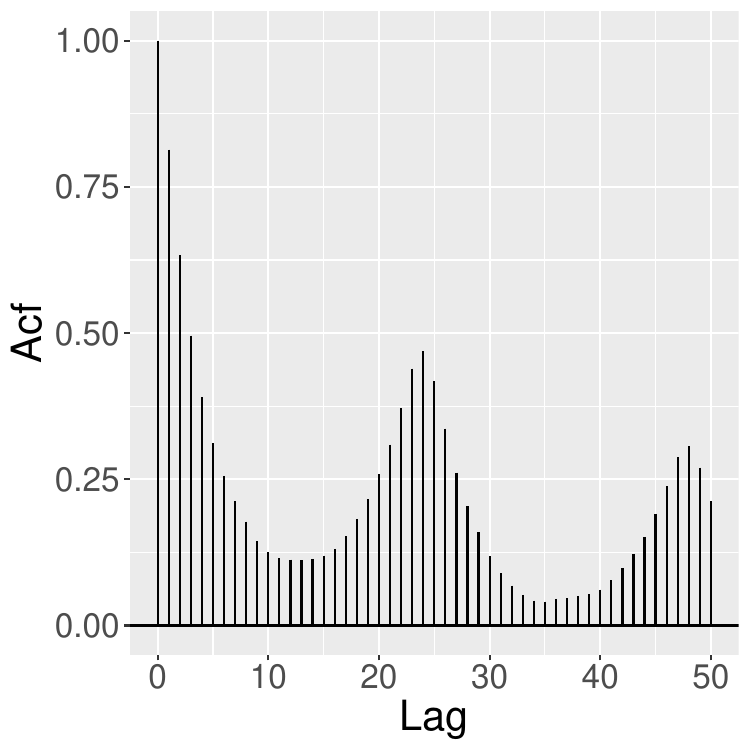}
  \caption{$\hat{\bY}^{\text{trend }}_t(\bs_N)$.}
\end{subfigure}
\vspace{5mm}
\begin{subfigure}{0.45\textwidth}
  \centering
\includegraphics[width=0.8\textwidth,]{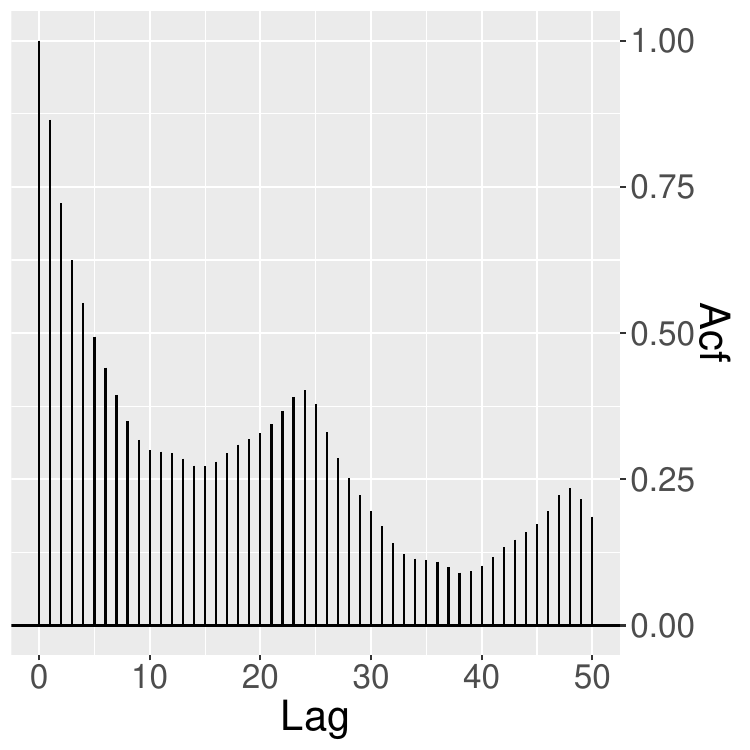}
  \caption{$\hat{\bY}^{\text{trend}}_t(\bs_S)$.}
\end{subfigure}
\vspace{5mm}
\begin{subfigure}{0.45\textwidth}
  \centering
  \includegraphics[width=0.8\textwidth,]{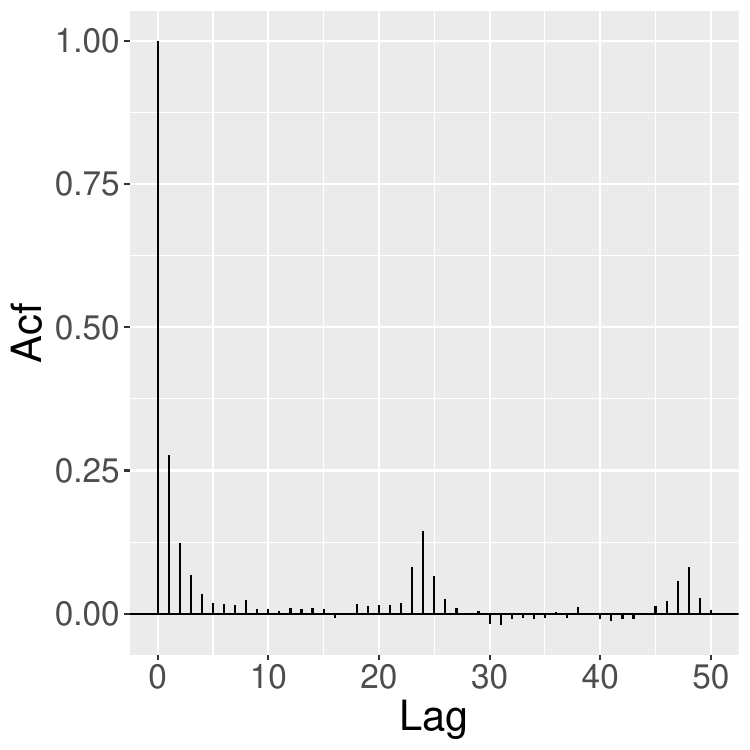}
    \caption{$\hat{\bY}^{\text{trend + time}}_t(\bs_N)$.}
\end{subfigure}
\begin{subfigure}{0.45\textwidth}
  \centering
  \includegraphics[width=0.8\textwidth,]{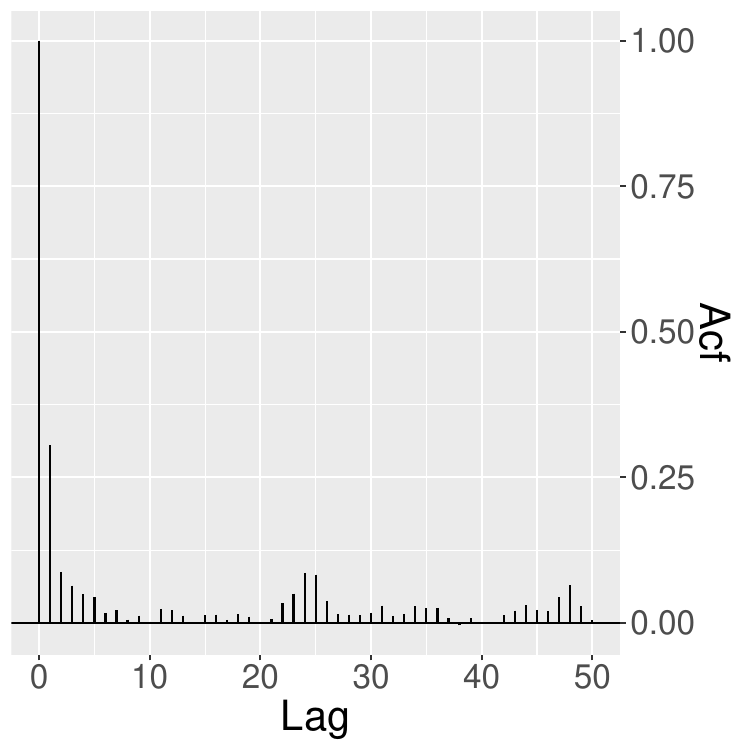}
 \caption{$\hat{\bY}^{\text{trend + time}}_t(\bs_S)$.}
\end{subfigure}
\caption{Auto-correlation function for $\hat{\bY}^{\text{trend}}_t(\bs_i)$ (first row) and $\hat{\bY}^{\text{trend+time}}_t(\bs_i)$ (second row), where the subscript refers to the N and S indicated as black crosses in Figure \ref{fig:wind_data}.}
\label{FS2_acf}
\end{figure}

\begin{figure}[H]
\centering
\begin{subfigure}{0.48\textwidth}
\includegraphics[width=1\textwidth]{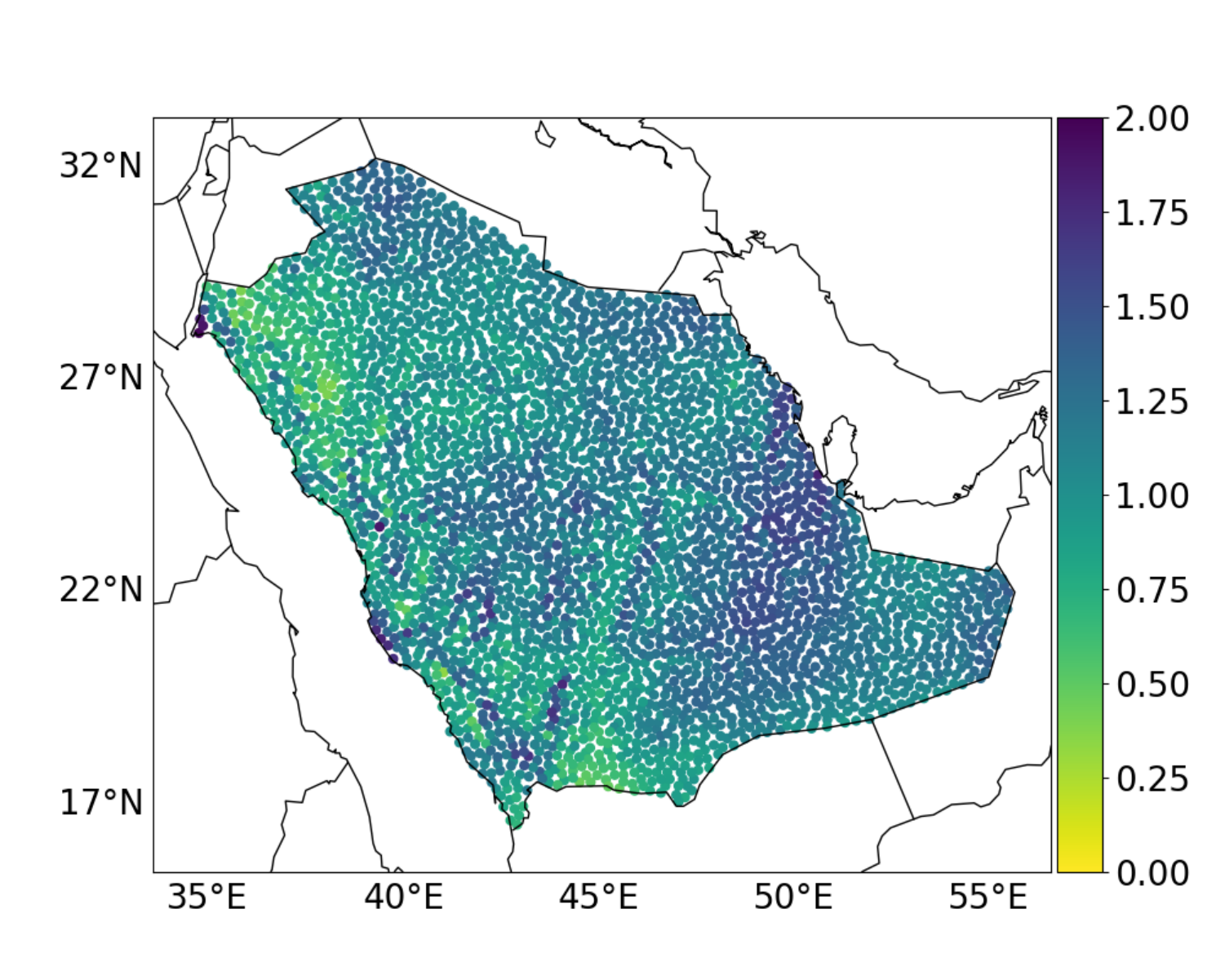}
\caption{One hour}
\end{subfigure}
\hspace{-8.5mm}
\begin{subfigure}{0.48\textwidth}
\includegraphics[width=1\textwidth]{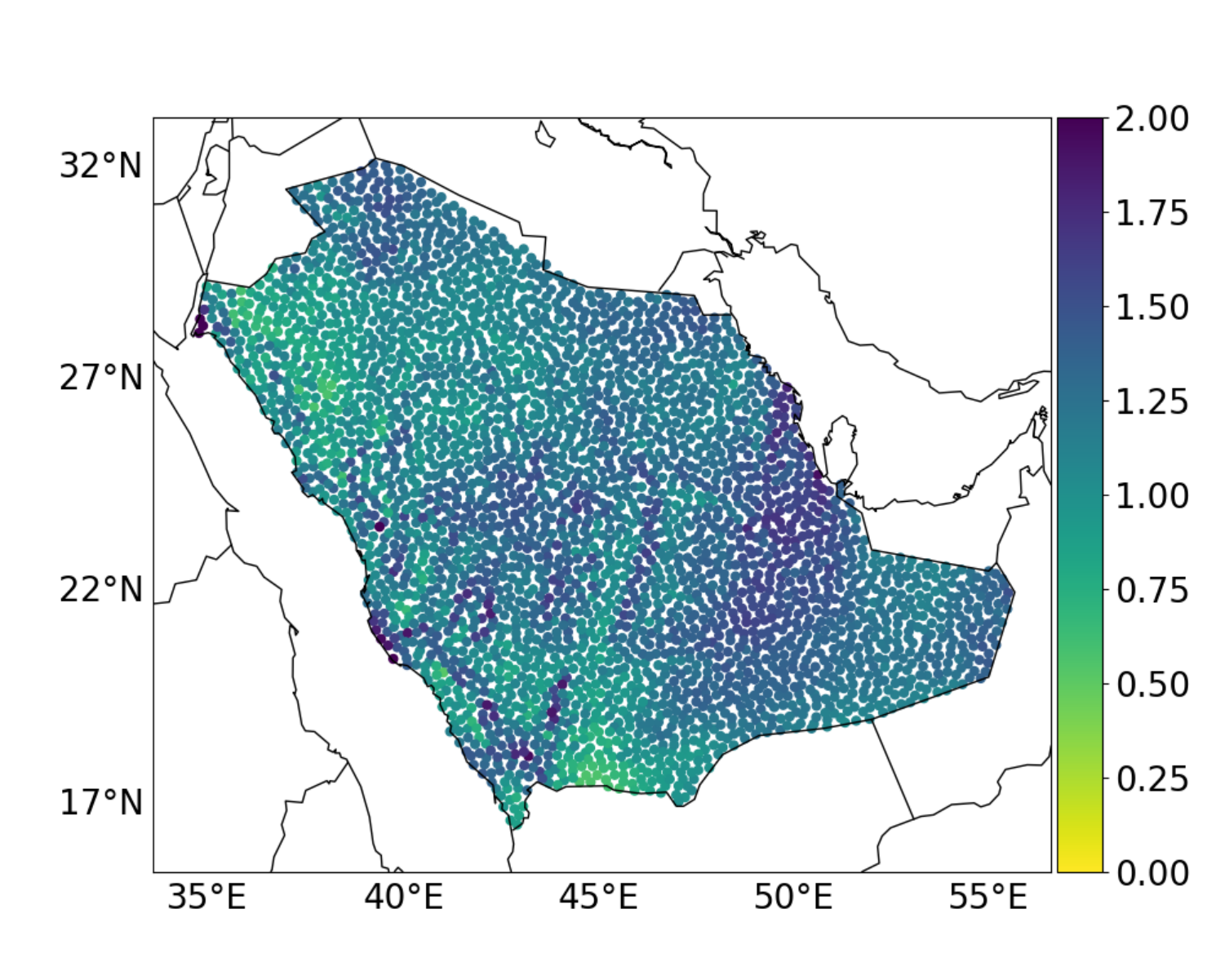}
\caption{Two hour}
\end{subfigure}
\hspace{-8.5mm}
\begin{subfigure}{0.48\textwidth}
\includegraphics[width=1\textwidth]{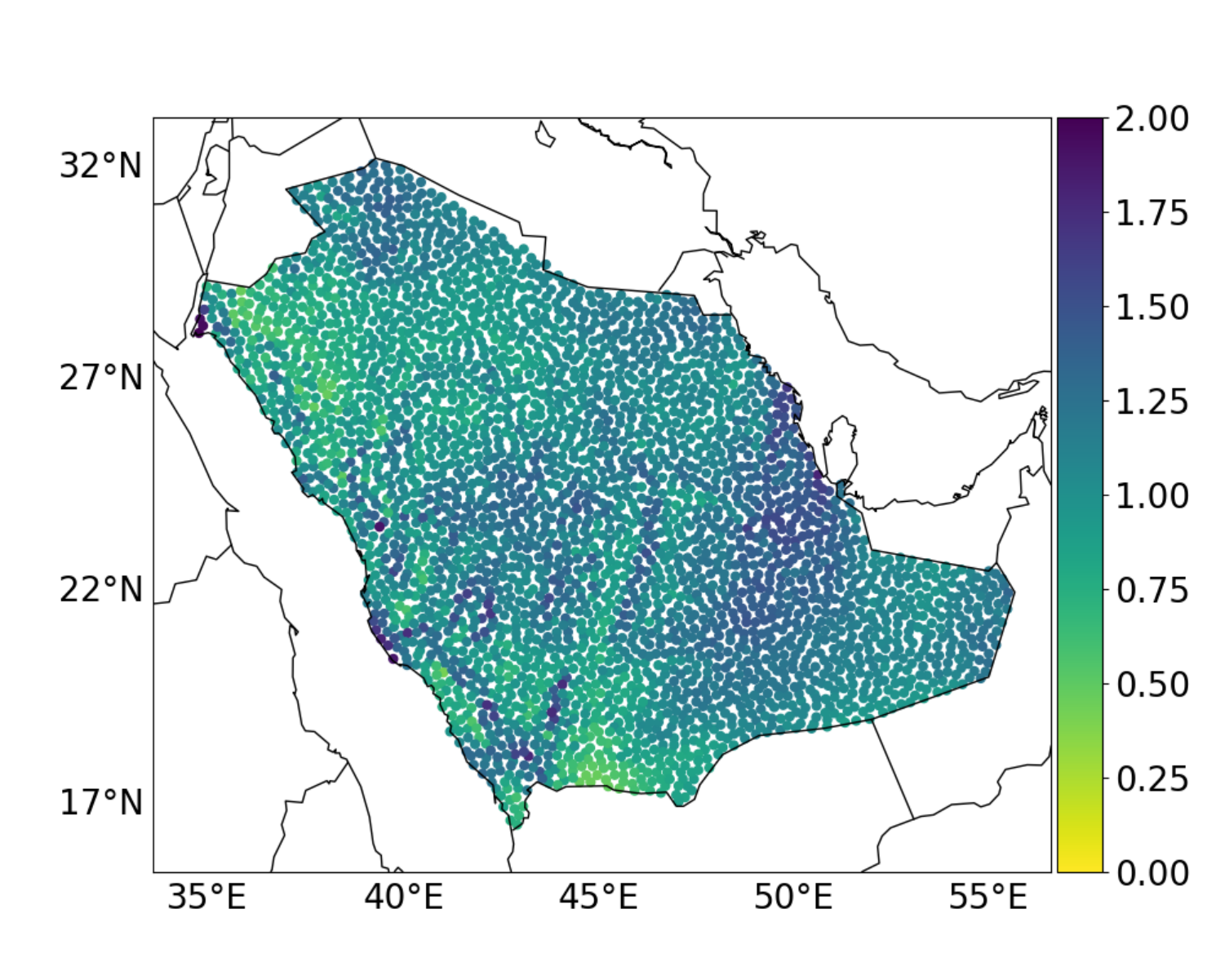}
\caption{Three hour}
\end{subfigure}
\caption{Maps of the differences between the $L^2$ norms of the auto-correlations (up to 50-hour) of $\hat{\bY}^{\text{trend}}_t(\bs_i)$ and $\hat{\bY}^{\text{trend+time}}_t(\bs_i), i=1, \ldots, n$. The medians (IQRs) of the $L^2$ norm differences are $1.110 (0.307)$, $1.212 (0.279)$, and $1.074 (0.270)$ for the three lead hours, respectively.}
\label{ACF_diff}
\end{figure}

Figure \ref{FS2_acf} reports the auto-correlation functions for the wind residuals ($\hat{Y}_t^{\text{trend}}(\bs)$) obtained from trend removal and the residuals resulted from trend and time dependence removal ($\hat{Y}_t^{\text{trend + time}}(\bs)$). Figure \ref{ACF_diff} shows the $L^2$ norm of the difference for the ACFs for these two residuals at first 50 time points at different lead times.

\newpage

\section*{Simulation Study}

In this section, two simulation studies are performed to asses different features in our model:
\begin{enumerate}
    \item the spatial structure of support points in terms of interpolation (kriging) accuracy;
    \item the temporal structure of the ESN in terms of forecasting skills.
\end{enumerate}

\subsection*{Spatial Structure}

We test the ability of our proposed support points approach to retain the spatial information. Since the temporal part is of no interest here, we drop the temporal index and work in a setting where the simulations are purely spatial. 

We simulate data from a bi-resolution spatial model \citep{castruccio2018scalable} over a domain $[0,1]^2$, which we split into four blocks $B_1, \ldots, B_4$ (see Figure \ref{FS5}) and define the following:
\begin{align*}
    \bY = & \bOmega \bH_{\text{loc}} +  (\bI - \bOmega)\bH_{\text{reg}},
\end{align*}
where $\bY = (Y(\mb{s}_1),\dots, Y(\mb{s}_n))^\top$ and $n = 3,200$ is the total number of locations. The two random effects $\bH_{\text{loc}}\in \Bbb{R}^n$ and $\bH_{\text{reg}}\in \Bbb{R}^n$ represent a local and regional effect, the former varying across every location (but according to a potentially different distribution across blocks), the latter being constant across each of the blocks. The matrix $\bOmega$ is diagonal with entries between 0 and 1 representing the relative weight of $\bH_{\text{loc}}$ with respect to $\bH_{\text{reg}}$. The latent processes $\bH_i \sim {\cal N}_n(\0,\bSigma_i), i \in \{\text{loc, reg}\}$ are independent Gaussian random fields modeled via a Mat\'ern correlation function \citep{ste99}:
\begin{align*}
       \text{corr}(Y(s_i),Y(s_j);\bsy{\theta}_i)&=\frac{1}{2^{\nu_i-1}\Gamma(\nu_i)}\left(\frac{d}{\beta_i}\right)^{\nu_i} {\cal K}_{\nu_i} \left(\frac{d}{\beta_i} \right), \quad &&\bsy{\theta}_i  = (\beta_i, \nu_i)^\top, i \in \{\text{loc, reg}\}.
\end{align*}
Here, $d=\|\mb{s}_i-\mb{s}_j\|$ is the Euclidean distance, ${\cal K}_{\nu_i}$ is the Bessel function of third kind of degree $\nu_i$, $\beta_i$ governs the spatial dependence range and $\nu_i$ dictates the smoothness of the spatial process. In this experiment, we simulate $\bY$ with $\bOmega = \diag(0.70\times\bsy{1}_{B_1},0.6 \times \bsy{1}_{B_2},0.9 \times\bsy{1}_{B_3},0.55\times\bsy{1}_{B_4})$, and we assume that $\btheta_{\text{loc}} = (0.1,0.5)^\top$ and $\btheta_{\text{reg}} = (0.03,1)^\top$ for $\bH_{\text{loc}}$ and $\bH_{\text{reg}}$, respectively. We choose a moderate dependence range and a low smoothness for the local effect $\bH_{\text{loc}}$, while for  $\bH_{\text{reg}}$ we choose a narrow dependence range and smooth spatial process that accounts for region-specific behavior. 

We then randomly divide the set of $n$ observations into two sets: $S_{\text{train}}$ and $S_{\text{test}}$, each one with 50\% of the original data (i.e., $|S_{\text{test}}|=n_{\text{test}} = 1,600$). Inference is then performed on $S_{\text{train}}$ using spatial reduction with the support points approach explained in Section \ref{sec:spred} (henceforth denoted as SP) and { another three commonly adopted approaches: \textit{SF}, \textit{Grid} and \textit{Rand}.
\textit{SF} denotes the space-filling approach implemented in the \textit{R} package \textit{fields} \citep{fields}}
\textit{Grid} indicates a spatial reduction technique that chooses the nearest neighbors of regularly spaced representative locations across the domain. \textit{Rand}, on the other hand, randomly selects the locations from the entire set with equal probability. We select $n_{\text{red}} = 100$ sites from the three approaches, and we use the data at these locations to perform spatial interpolation on $\bY_{\text{test}}$ (the testing set) with the SPDE model from equation \eqref{eq:SPDE}. We perform $n_{\text{sim}}=100$ simulations using two different sampling schemes as well as different realizations of the bi-resolution model. The first sampling scheme, which we denote as \textit{chessboard}, is a block-wise preferential pattern in which the domain is partitioned into evenly spaced blocks and locations are densely populated in certain blocks but sparsely in the others (see Figure \ref{FA_L}). The second one, which we denote as \textit{ray}, preferentially samples locations in a block, and within that block concentrates points within a subregion (see Figure \ref{FA_R}). Technical details on both sampling schemes are provided in the Supplementary Material. As a measure of comparison for { \textit{SF}}, \textit{SP}, \textit{Grid} and \textit{Rand}, for each simulation we compute the Mean Squared Prediction Error (MSPE), defined as $1/n_{\text{test}}\sum_{i \in S_{\text{test}}} \{ \bY(\mb{s}_i) - \hat{\bY}(\mb{s}_i)\}^2$, where $\hat{\bY}(\mb{s}_i)$ is the fitted value according to the different sampling methods using our SPDE approach for interpolation. In other words, once inference is performed on the training set according to each of the three sampling approaches, we compute the difference between the true value and the predicted one in the testing set.

\begin{figure}[t!]
\centering
\begin{subfigure}{0.45\textwidth}
\centering
\includegraphics[width=1\textwidth,]{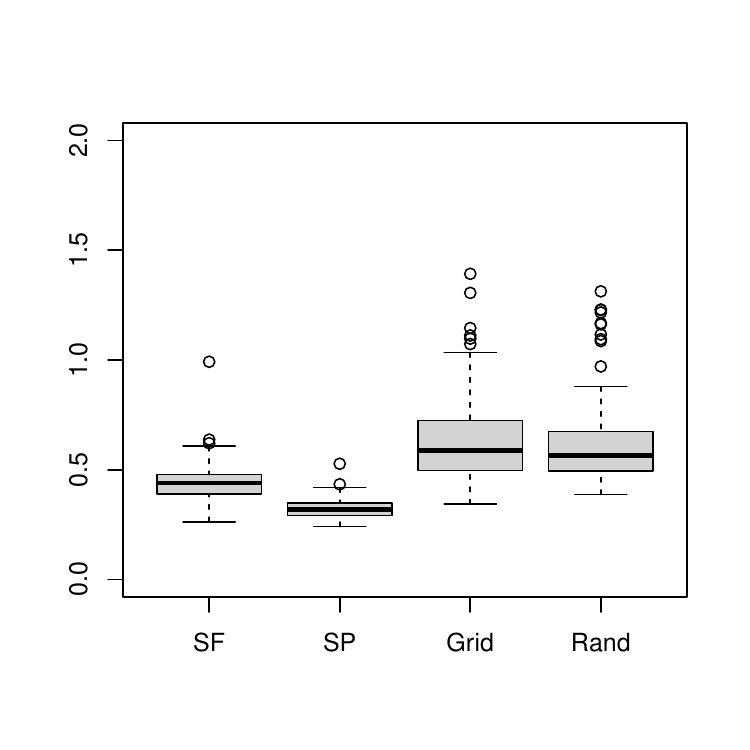}
\vspace{-10mm}
\caption{{ Chessboard.}}  \label{fig:siml}
\end{subfigure}
\begin{subfigure}{0.45\textwidth}
\centering
\includegraphics[width=1\textwidth,]{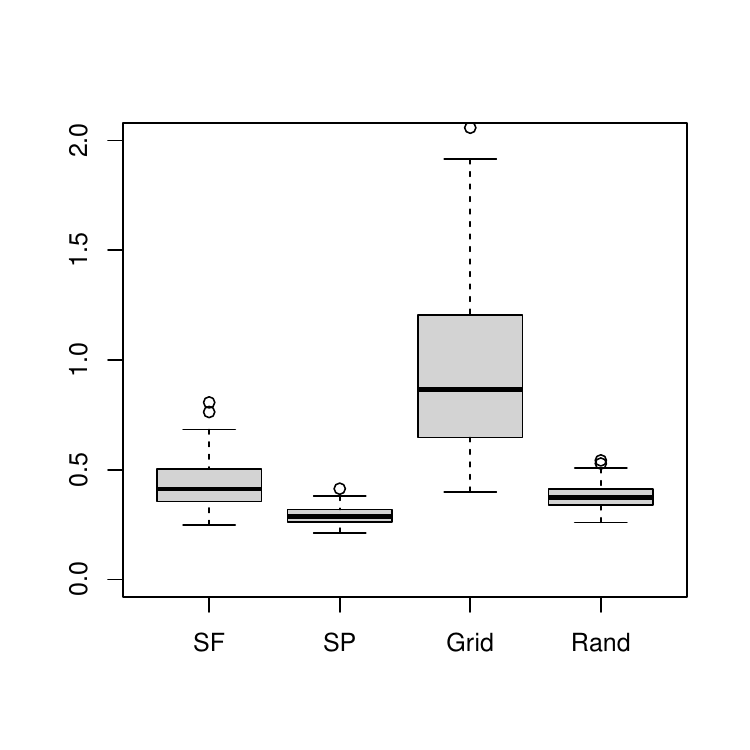}
\vspace{-10mm}
\caption{{ Ray.}}  \label{fig:simr}
\end{subfigure}
\caption{Boxplots (across simulations) of the MSPEs of the spatial interpolation over the test locations calculated from the chosen locations through \textit{SF}, \textit{SP}, \textit{Grid}, and \textit{Rand} on a (a) chessboard (see Figure \ref{FA_L}) and (b) ray domain (see Figure \ref{FA_R}).}
  \label{fig:sim1}
\end{figure}

\begin{figure}[H]
\centering
\begin{subfigure}{0.24\textwidth}
\includegraphics[width=1\textwidth]{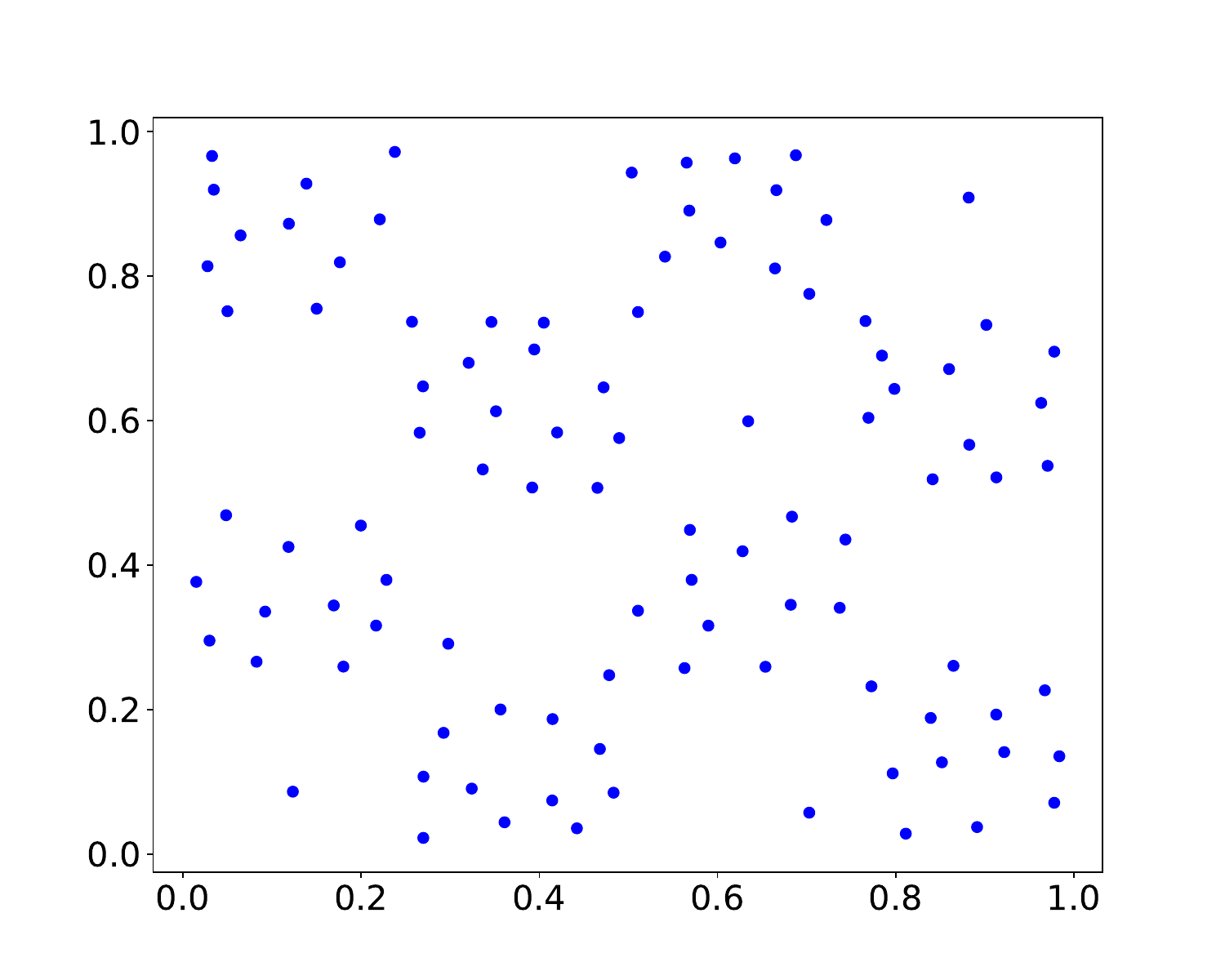}
\caption{Chessboard-SP}
\label{sp}
\end{subfigure}
\hspace{-5mm}
\begin{subfigure}{0.24\textwidth}
\includegraphics[width=1\textwidth]{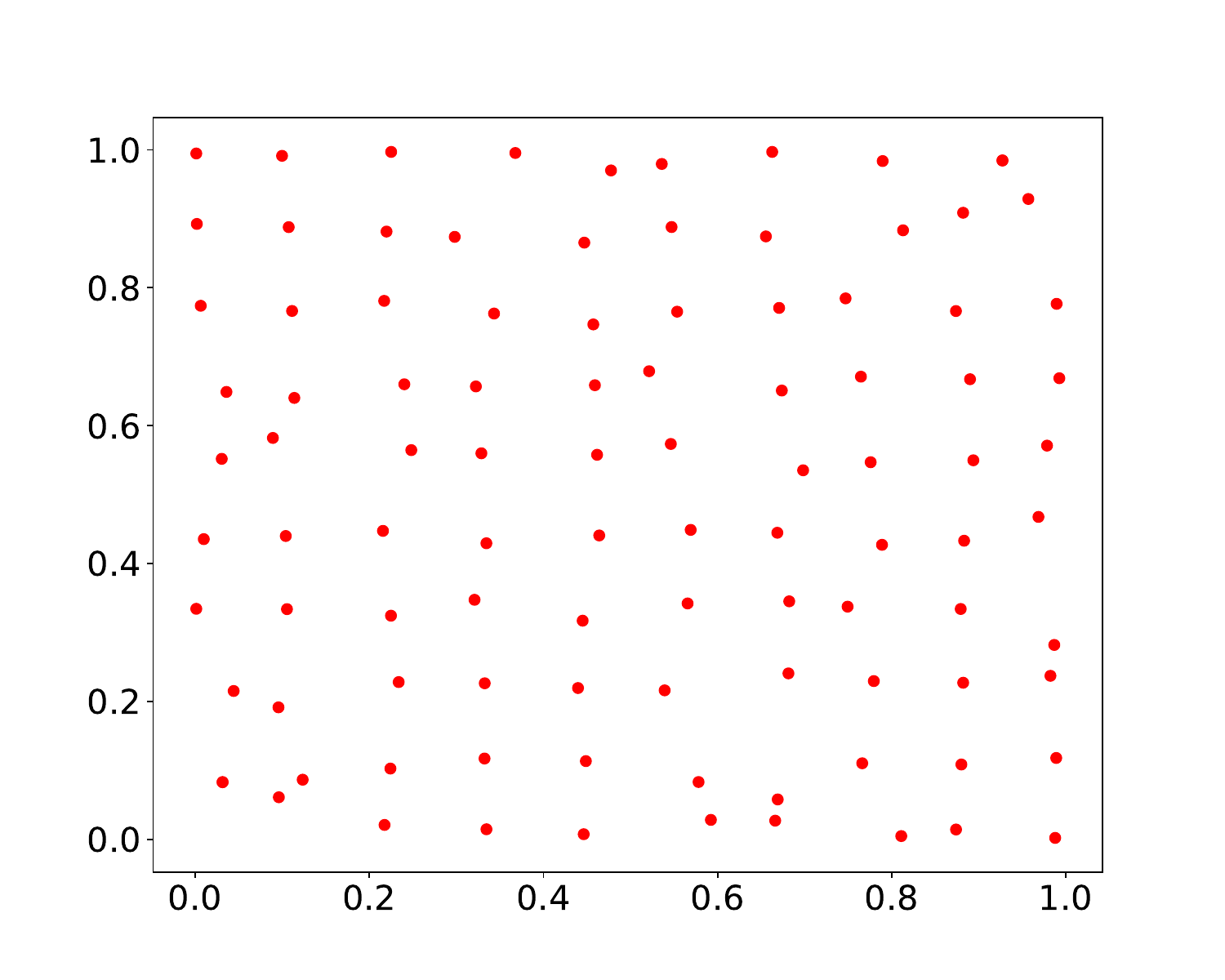}
\caption{Chessboard-Grid}
\label{grid}
\end{subfigure}
\hspace{-5mm}
\begin{subfigure}{0.24\textwidth}
\includegraphics[width=1\textwidth]{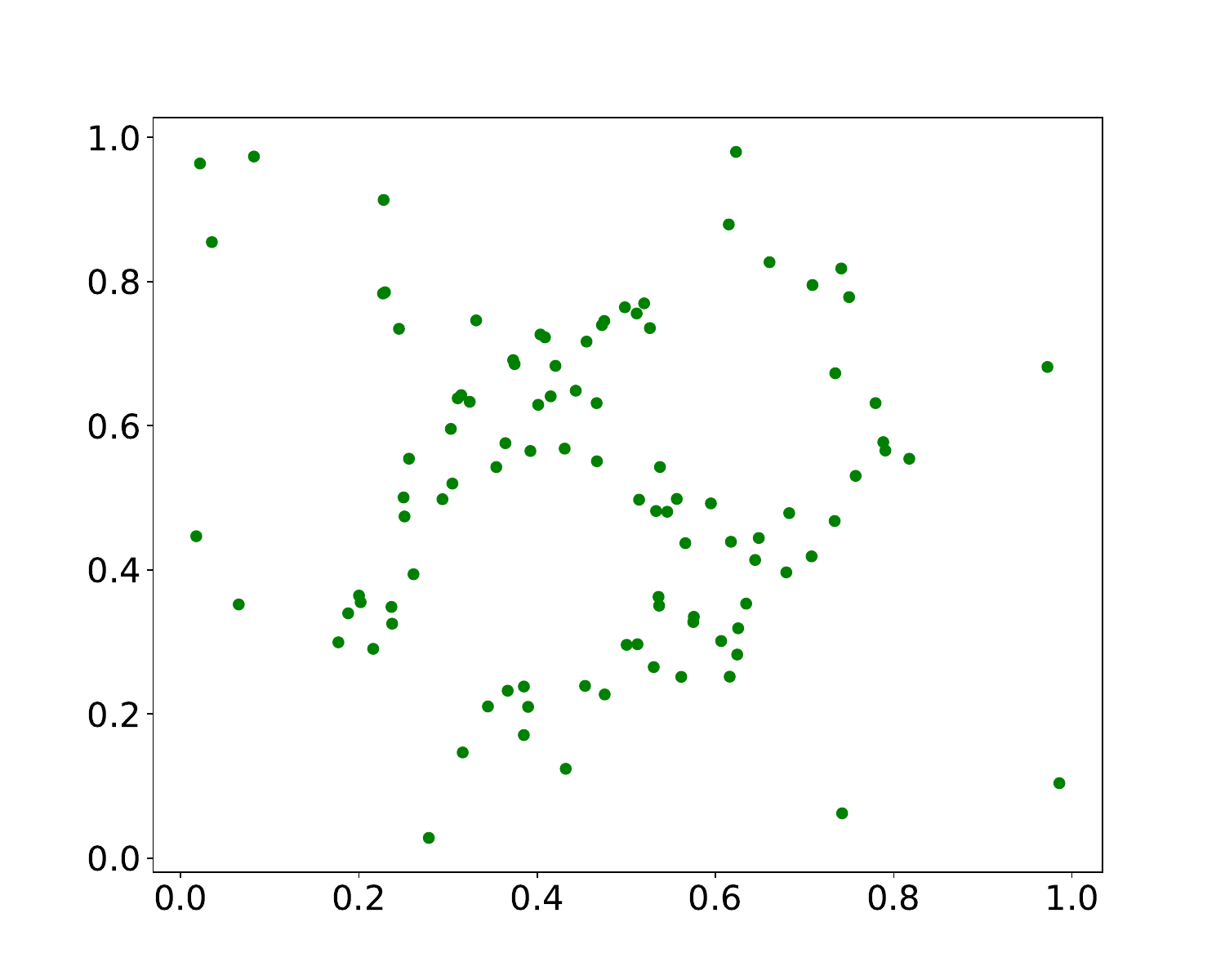}
\caption{Chessboard-Rand}
\label{rand}
\end{subfigure}
\hspace{-5mm}
\begin{subfigure}{0.24\textwidth}
\includegraphics[width=1\textwidth]{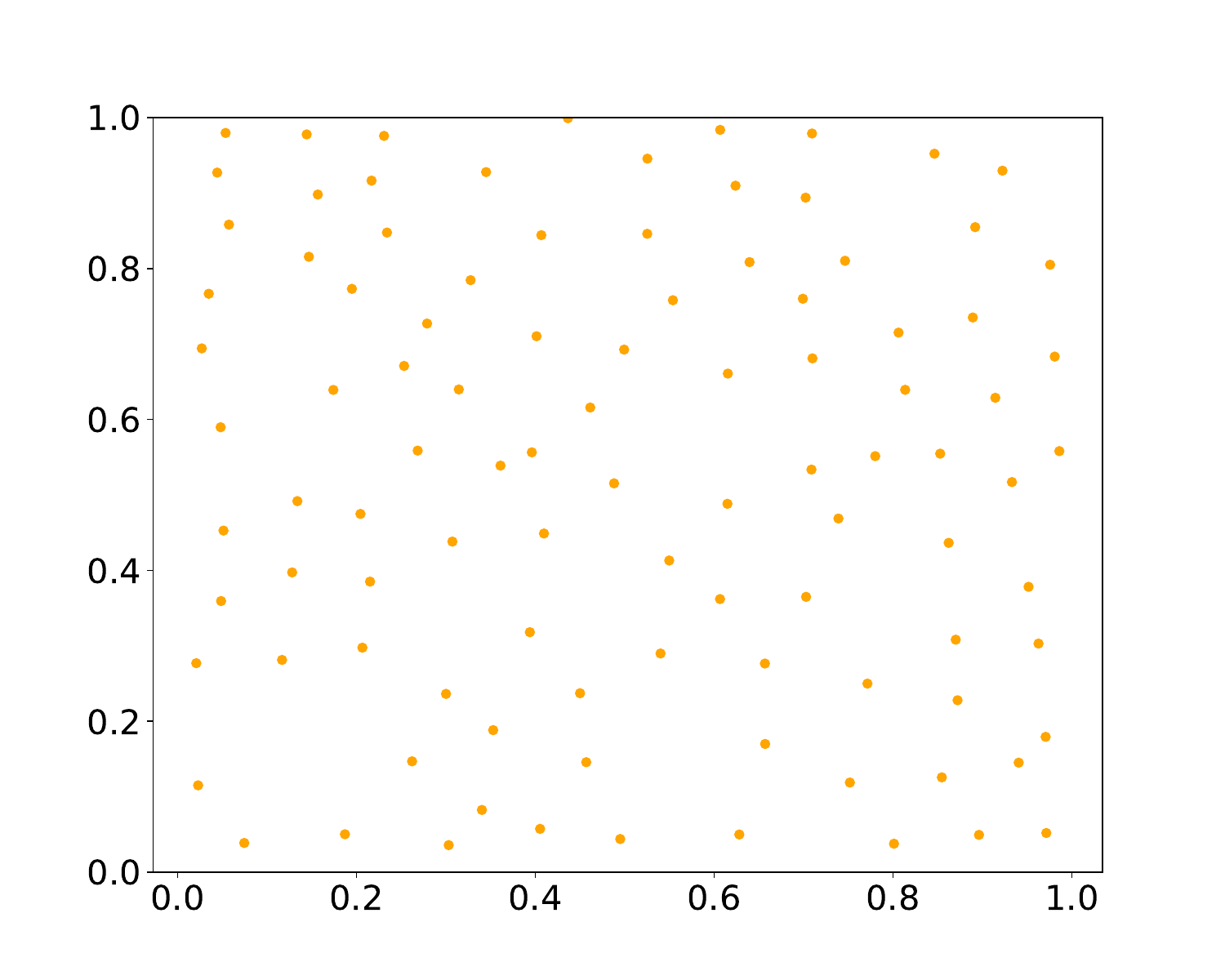}
\caption{Chessboard-SF}
\label{SF}
\end{subfigure}
\begin{subfigure}{0.24\textwidth}
\includegraphics[width=1\textwidth]{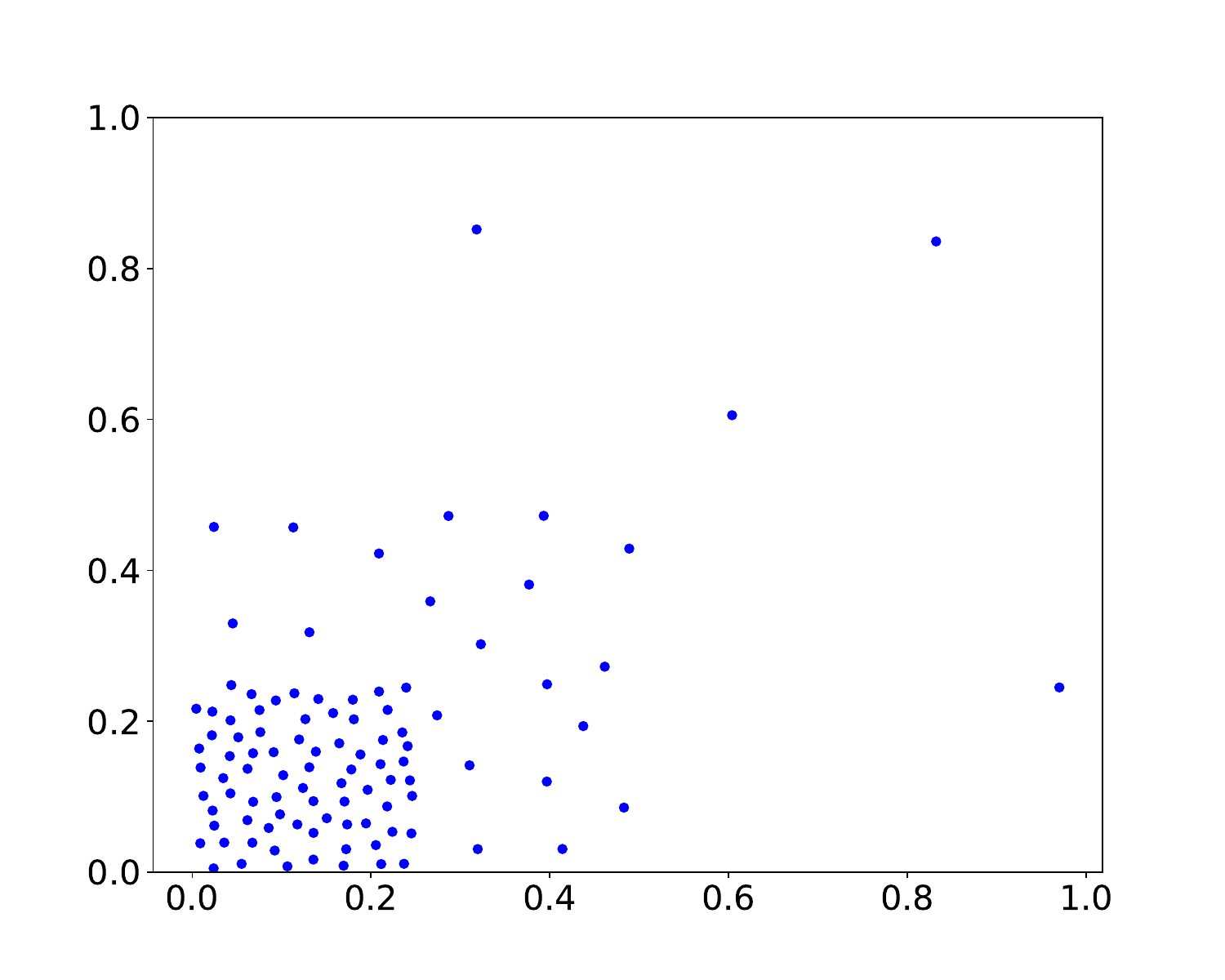}
\caption{Ray-SP}
\label{sp_u}
\end{subfigure}
\hspace{-5mm}
\begin{subfigure}{0.24\textwidth}
\includegraphics[width=1\textwidth]{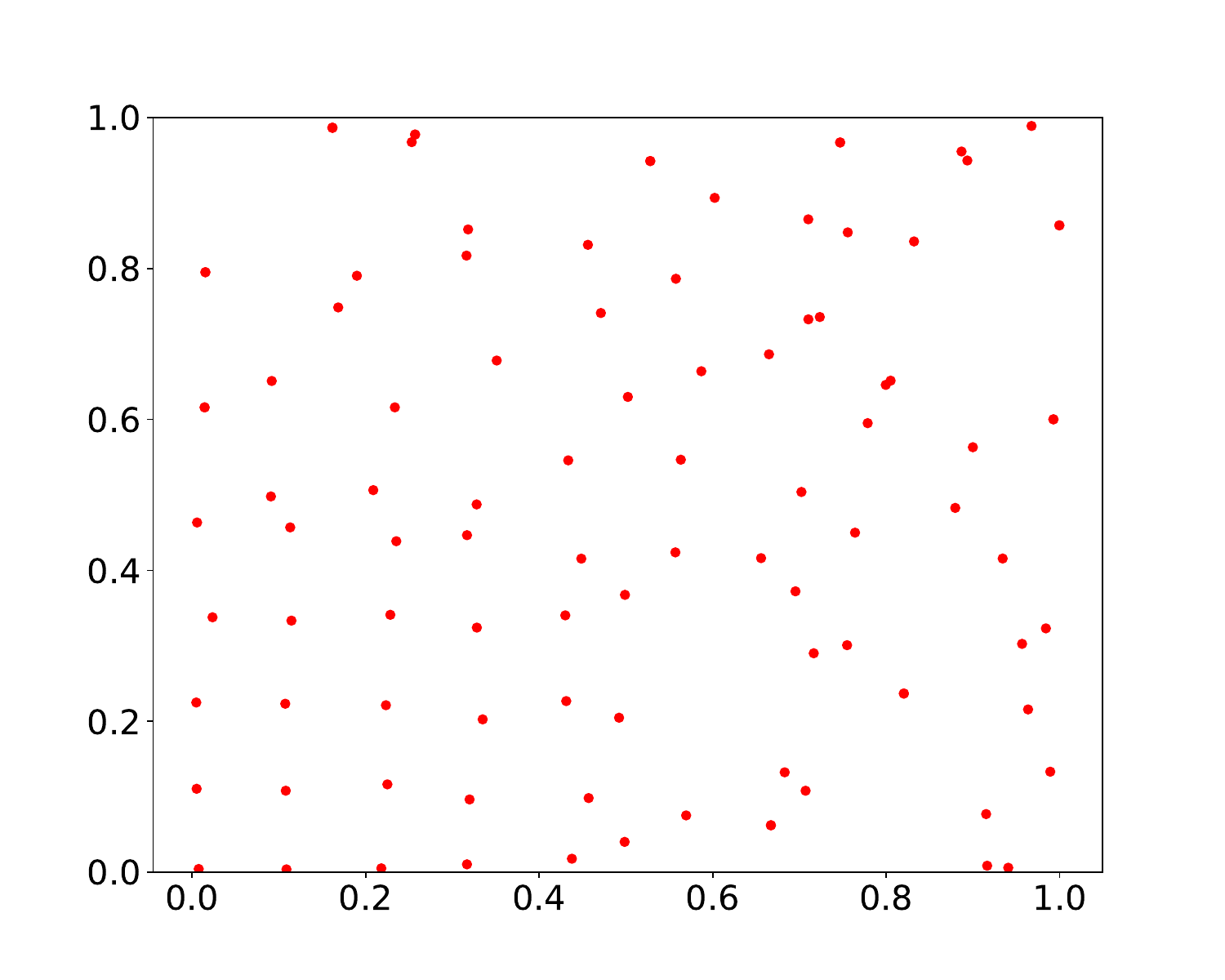}
\caption{Ray-Grid}
\label{grid_u}
\end{subfigure}
\hspace{-5mm}
\begin{subfigure}{0.24\textwidth}
\includegraphics[width=1\textwidth]{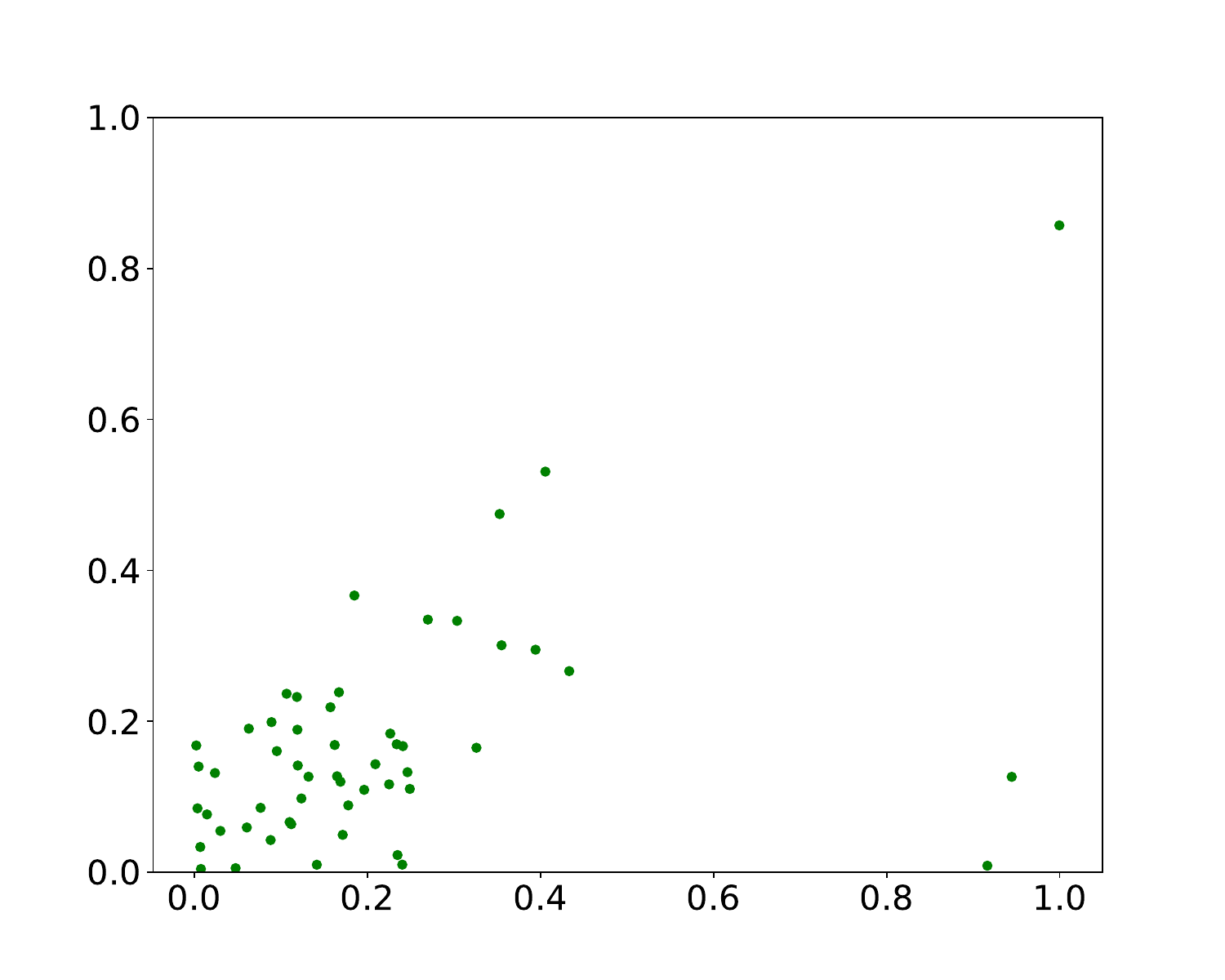}
\caption{Ray-Rand}
\label{rand_u}
\end{subfigure}
\hspace{-5mm}
\begin{subfigure}{0.24\textwidth}
\includegraphics[width=1\textwidth]{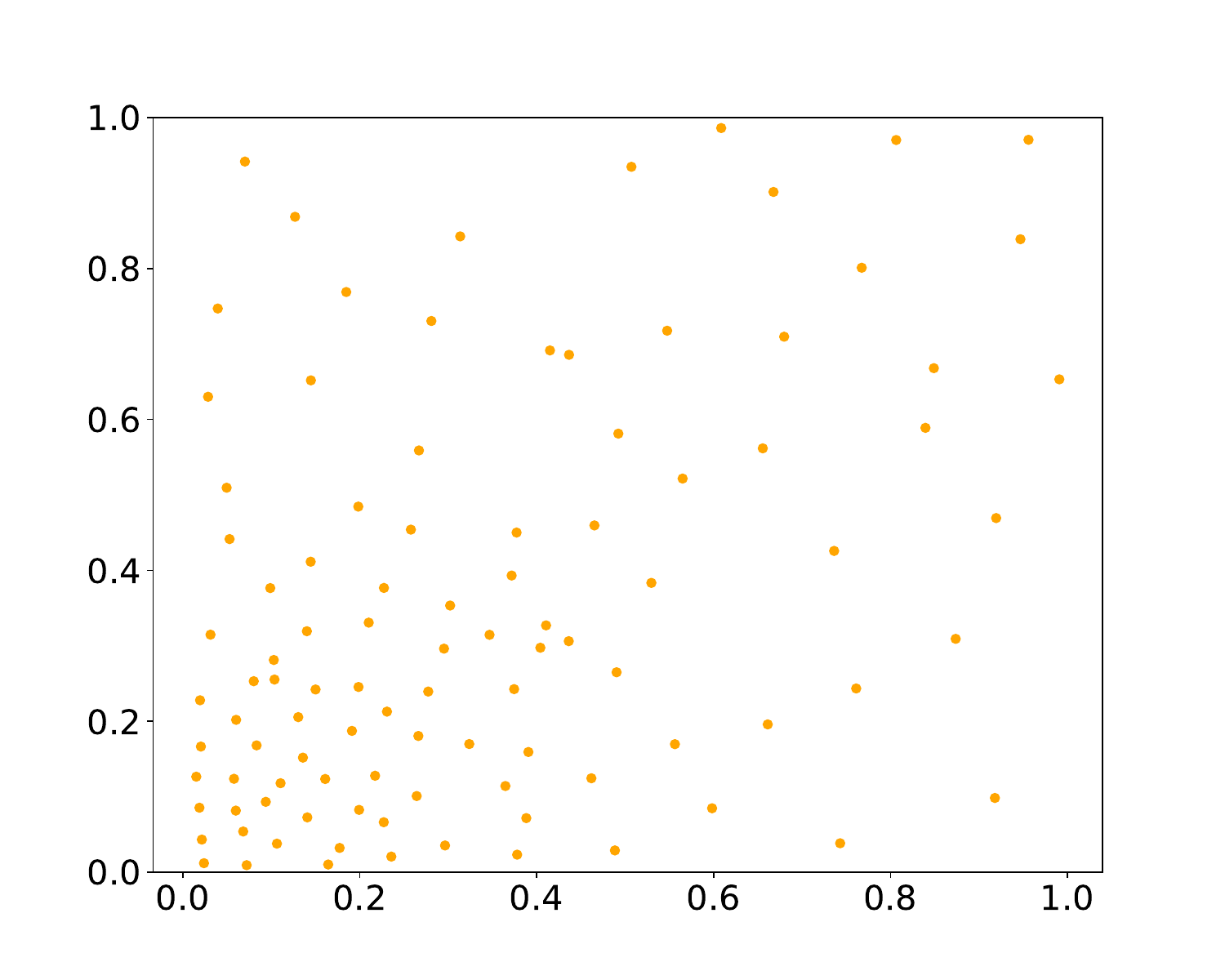}
\caption{Ray-SF}
\label{SF_u}
\end{subfigure}
\caption{{ Example of the SP, Grid, Rand, and SF spatial reductions on a (a-d) \textit{chessboard} and (e-f) \textit{ray} sampling scheme.}}
\label{r_u}
\end{figure}

Figure \ref{fig:sim1} shows the boxplots (across the $n_{\text{sim}}$ simulations) of the MPSE for \textit{SP}, \textit{Grid} and \textit{Rand}. In both cases, it is readily apparent how \textit{SP} is able to achieve more accurate spatial interpolations. Indeed, in the case of the \textit{chessboard} scheme, \textit{SP} achieves a median MSPE of 0.317 (IQR 0.054) compared to 0.596 (0.222) of \textit{Grid} and 0.572 (0.212) of \textit{Rand}. This is also visually apparent when observing the chosen locations by SP, Grid and Rand in a simulation in Figure \ref{r_u}(a-c). Indeed, \textit{SP} is able to capture the clustering effect of the locations. \textit{Rand} is less efficient but is still able to capture the pattern to a limited degree because locations from a densely populated clusters are more likely to be selected. \textit{Grid}, on the other hand, leads to the most inefficient reconstructions as regularly spaced locations are incapable of acknowledging local clusters. Similarly, results from the \textit{ray} sampling scheme in Figure \ref{fig:sim1}b show a median MSPE of 0.287 (IQR 0.055) for \textit{SP} compared to 0.865 (0.554) of \textit{Grid} and 0.372 (0.072) of \textit{Rand}, again confirming the superior efficiency of \textit{SP}. Unlike the \textit{chessboard} case involving multiple densely populated patches, the \textit{ray} case contains blocks, in which the population density decades sharply from the lower left (see Figure \ref{FA_R}). In this instance, \textit{Rand} is also somewhat able to capture the underlying pattern although not as efficiently as \textit{SP}.

{ As for the SF approach, locations generated using this method are less representative of the geometry of the original chessboard compared with the SP approach; see Figure~\ref{SF}. The chosen locations are close to be uniformly distributed. In case of a ray, although the chosen locations can better mimic the shape of the original data compared with Grid but concentrates more weights on the supposedly sparse regions compared with SP and Rand; see Figure~\ref{SF_u}. The prediction efficiency of SF (Chessboard: median MSPE - 0.426, IQR - 0.099; Ray: median - 0.412, IQR - 0.147) is also lower compared to SP due to the less expressiveness of geographical geometry, which is shown in Figure~\ref{fig:sim1}.}

\subsection*{Temporal Structure}

We now compare the forecasting skills of our B-ESN approach against popular alternative machine learning methods such as the Long-Short-Term Memory (LSTM, \cite{hochreiter1997long}) and Gated Recurrent Unit (GRU, \cite{chung2014empirical}). Over the entire experiment, we use the Adaptive Moment Estimation (ADAM, \cite{kingma2014adam}) approach to estimate the parameters, and we tuned the networks with the best combination consisting of 80 hidden states, 2 hidden layer and 20\% dropout rate (see Figure \ref{TD96} in the Supplementary Material). We also benchmark our proposed approach against other statistical models such as Vector AutoRegressive (VAR, \cite{lutkepohl2013vector}) with one lag and Lattice Kriging (LAT, \cite{nyc15}), as well as a persistence (PER) approach, a na\"ive forecasting assuming all future observations are equal to the present. 

We simulate data from the Lorenz-96 model \citep{lorenz1996predictability}, a well-known dynamical system defined as:
\begin{align*}
    \frac{\partial Y_t(\mb{s}_i)}{\partial t} = \{Y_t(\mb{s}_{i+1}) - Y_t(\mb{s}_{i-2})\} Y_t(\mb{s}_{i-1}) - Y_t(\mb{s}_i) + F_t, \quad i = 1,\dots, n.
\end{align*}
We discretize the equation and simulate $n = 81$ vector elements, and a time series with $T = 1,000$ time points, with the first $T_{\text{train}} = 800$ used as training and the remaining $T_{\text{test}} = 200$ as testing. We set $F_t = 4.5$ to obtain a moderate absolute correlation (0.42). Various forcing constants produce differing absolute correlations among the vector elements but have overall small impact on the model forecasting efficiency \citep{bon23}. A total of $n_{\text{sim}}=100$ simulations are performed. 

\begin{figure}[b!]
\centering
\begin{subfigure}{0.32\textwidth}
\centering
\includegraphics[width=1\textwidth,]{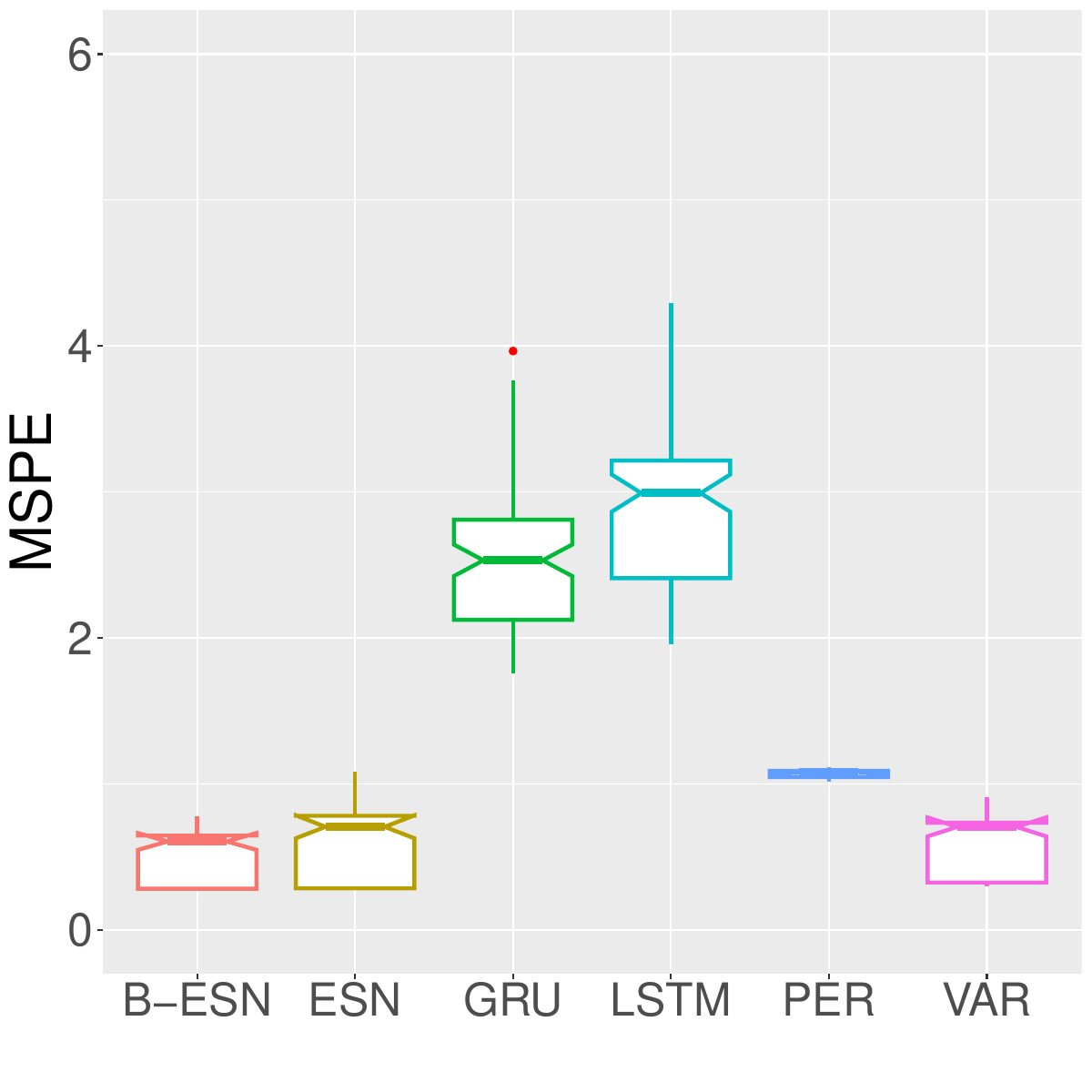}
\caption{One hour.}
\end{subfigure}
\begin{subfigure}{0.32\textwidth}
\centering
\includegraphics[width=1\textwidth,]{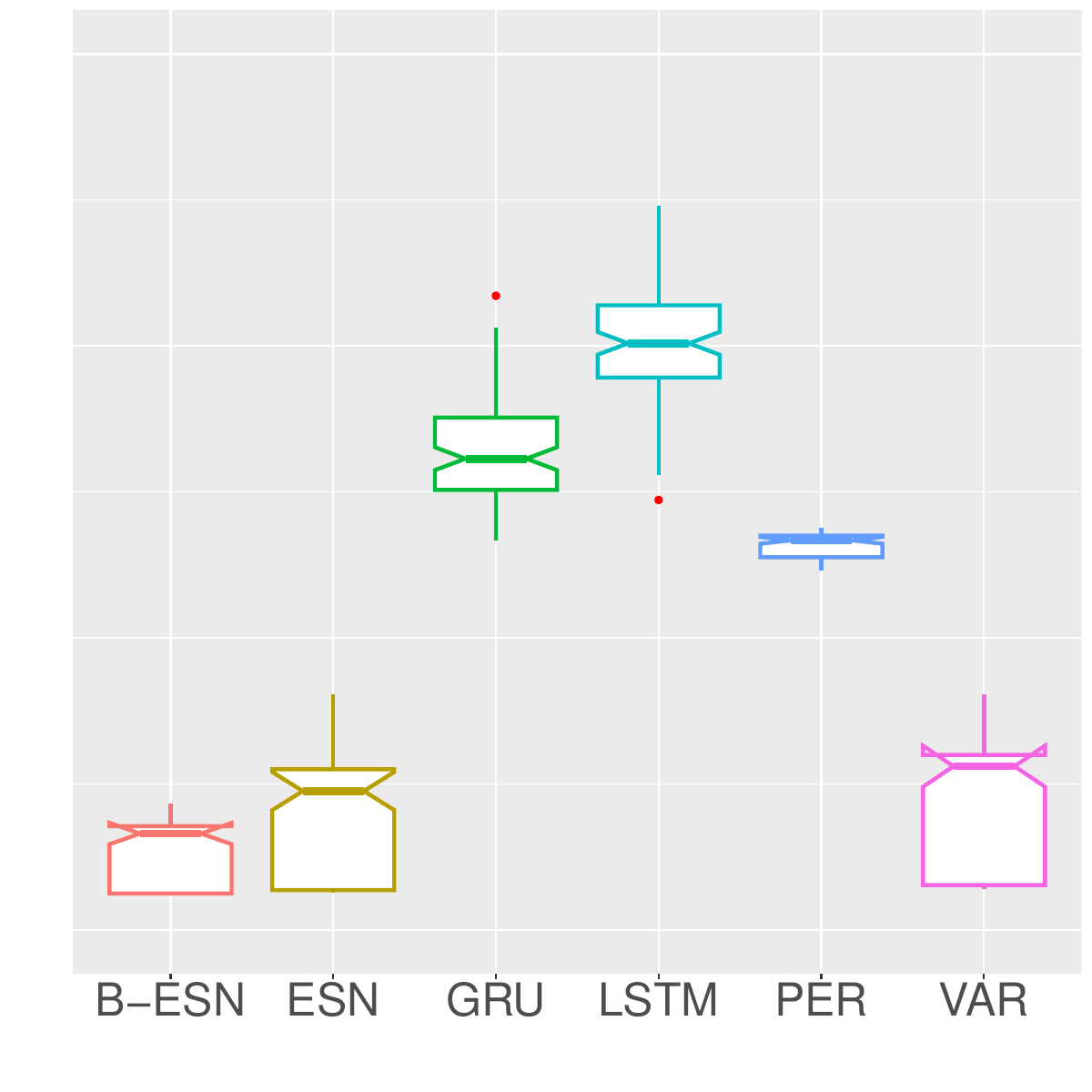}
\caption{Two hours.}
\end{subfigure}
\begin{subfigure}{0.32\textwidth}
\centering
\includegraphics[width=1\textwidth,]{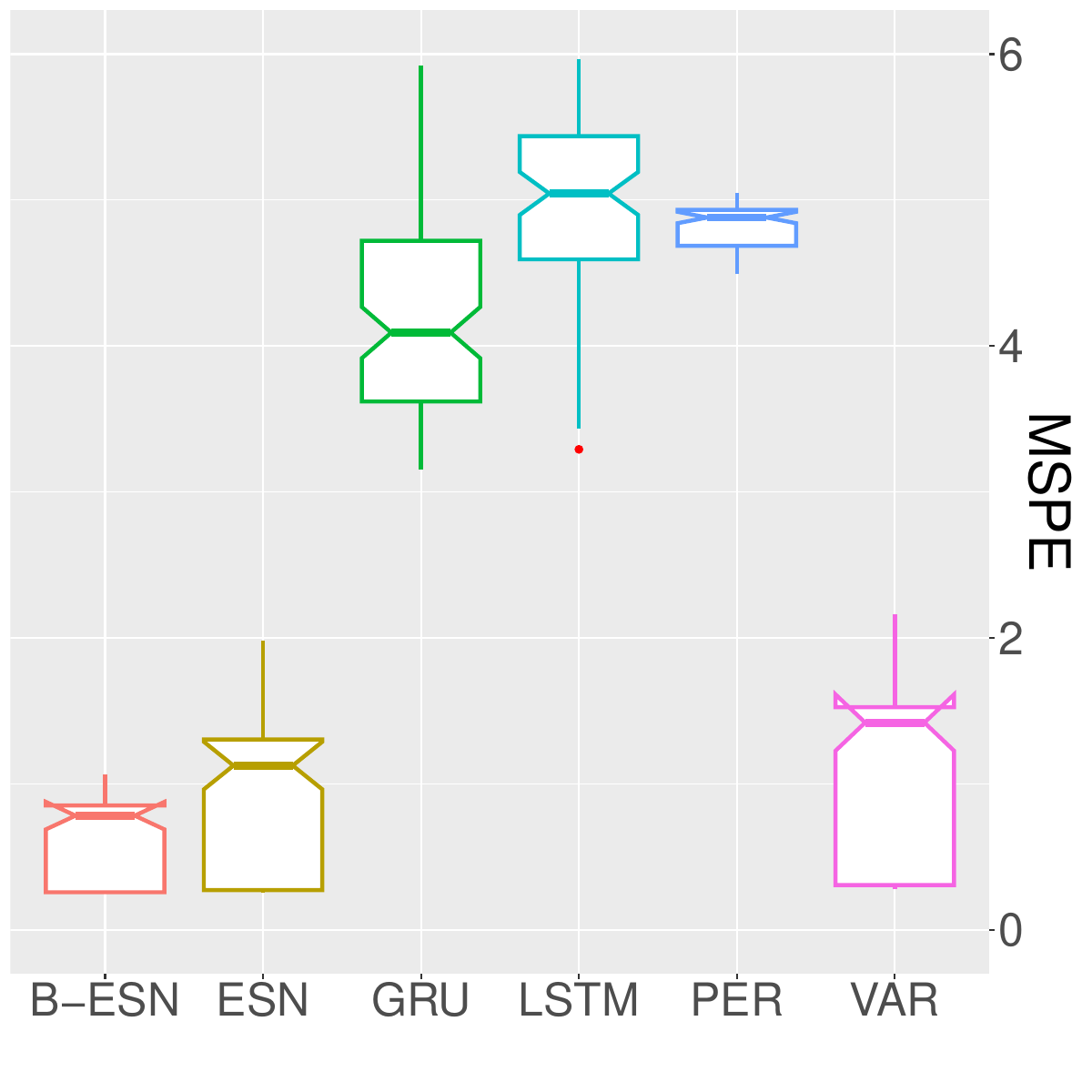}
\caption{Three hours.}
\end{subfigure}
\caption{Boxplot of the MSPEs (averaged across $n$ vector elements and $T_{\text{test}}$ time points) across forecasting lead times using our proposed B-ESN against ESN, GRU, LSTM, PER, and VAR models. The boxplot is computed by aggregating results across the $n_{\text{sim}}$ simulations. }
\label{fig:Lorenz96}
\end{figure}

We assess the forecasting ability of the different models through the MSPEs, defined here as $1/(n \cdot T_{\text{test}}) \sum_{t = T_{\text{train}} + 1}^{T}\sum_{i=1}^n \{\bY_t(\mb{s}_i) - \hat{\bY}_t(\mb{s}_i)\}^2$ where $\hat{Y}_t(\mb{s}_i)$ is the forecast according to one of the three sampling schemes. Figure \ref{fig:Lorenz96} indicates that our B-ESN approach has better predictive skills compared with the chosen machine learning and statistical models. Indeed, the MSPE of B-ESN is lower and the distributions (across simulations) is more stable, reflected through the smaller median values and narrow IQRs across varying lead times. 

LSTM and GRU are recurrent neural network models and both perform significantly worse then B-ESN. Figure \ref{TD96} provides additional evidence as to why this is the case, by indicating how the MSPE changes as a function of the \textit{epochs} (i.e., iterations) in the minimization algorithm. It is readily apparent how for a relatively short time series such as the one in this simulation study, LSTM and GRU achieve a stable value after less than a hundred epochs, with large MSPEs on the testing data compared with B-ESN, ESN, and VAR. VAR has overall competitive forecasts compared to B-ESN and ESN but with declining performances for long-lead forecasts. Forecasts obtained with PER are stable across the vector elements but suboptimal, and are progressively worse with long-lead forecasts.  Lastly, the ESN approach is expected to be less efficient than B-ESN (as discussed in Section \ref{batch-update} and \ref{sec:comp}) because it does not update the weight parameters at every lead time. 

\subsection*{Simulation 1: Spatial Structure}

\begin{figure}[H]
\centering
 \begin{subfigure}{0.45\textwidth}
\includegraphics[width=1\textwidth]{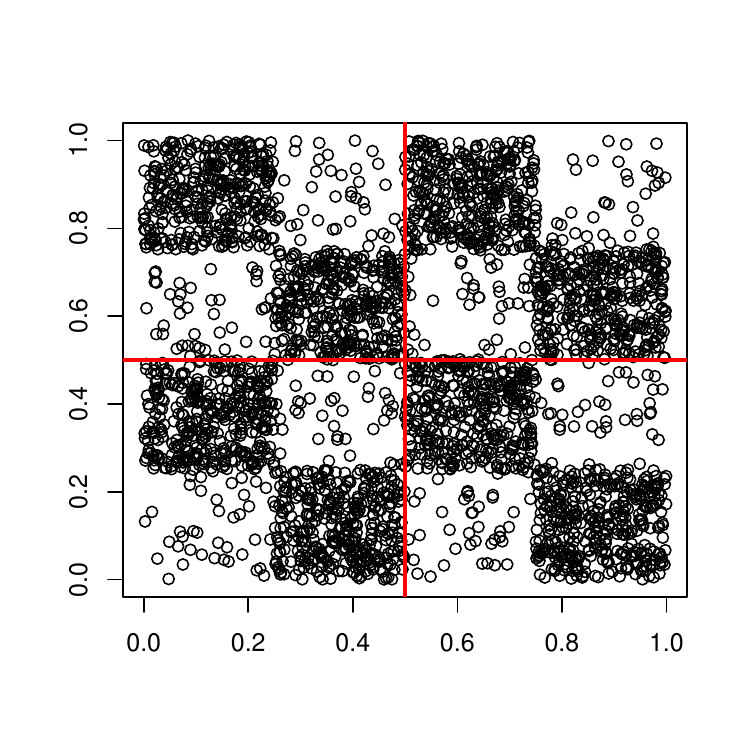}
\caption{Chessboard.}
\label{FA_L}
\end{subfigure}
  \begin{subfigure}{0.45\textwidth}
\includegraphics[width=1\textwidth]{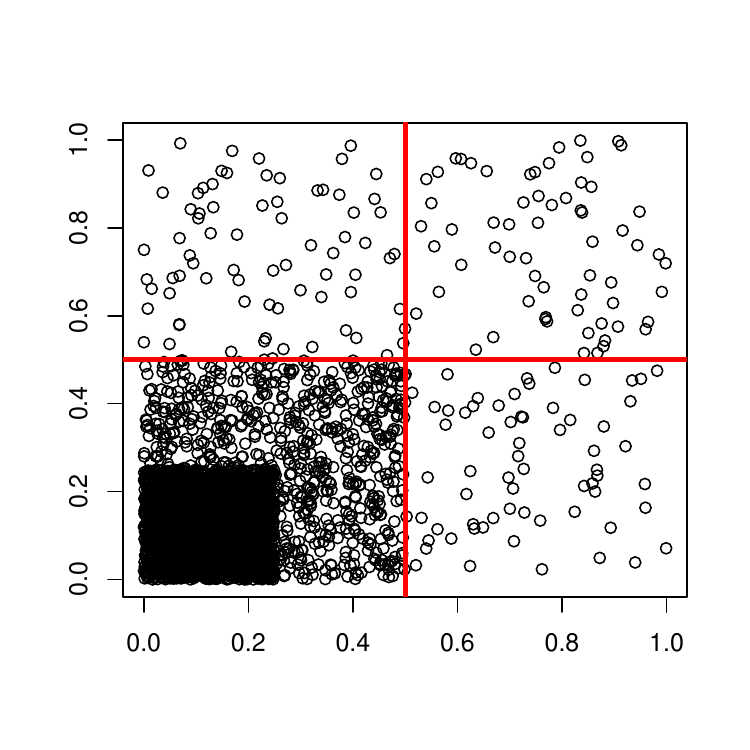}
\caption{Ray.}
\label{FA_R}
\end{subfigure}
  \caption{Sampling schemes for the spatial simulation study 1. Panel (a) represents a \textit{chessboard} sampling scheme while panel (b) represents a \textit{ray} sampling scheme. The red lines denote the blocks used to simulate the regional effect in the bi-resolution model.}
  \label{FS5}
\end{figure}

Figure \ref{FS5} shows the patterns of the \textit{chessboard} and \textit{ray} sampling scheme. For \textit{chessboard}, we first partition the unit square into 16 identical blocks. Out of the 16 blocks, we pick 8 dense blocks and 8 sparse blocks, in which the simulated locations are 9 times more likely to fall in the dense blocks than the sparse ones. For the \textit{ray} sampling scheme, we first set the boundaries of the three regions, which in Figure \ref{FA_R} are increasingly large squares. Second, we keep the number of sampled locations the same within each region. As a result, the locations will be more concentrated on the smaller regions.

\subsection*{Simulation 2: Temporal Structure}

\begin{figure}[H]
\centering
\begin{subfigure}{0.3\textwidth}
  \centering
  \includegraphics[width=1\textwidth,]{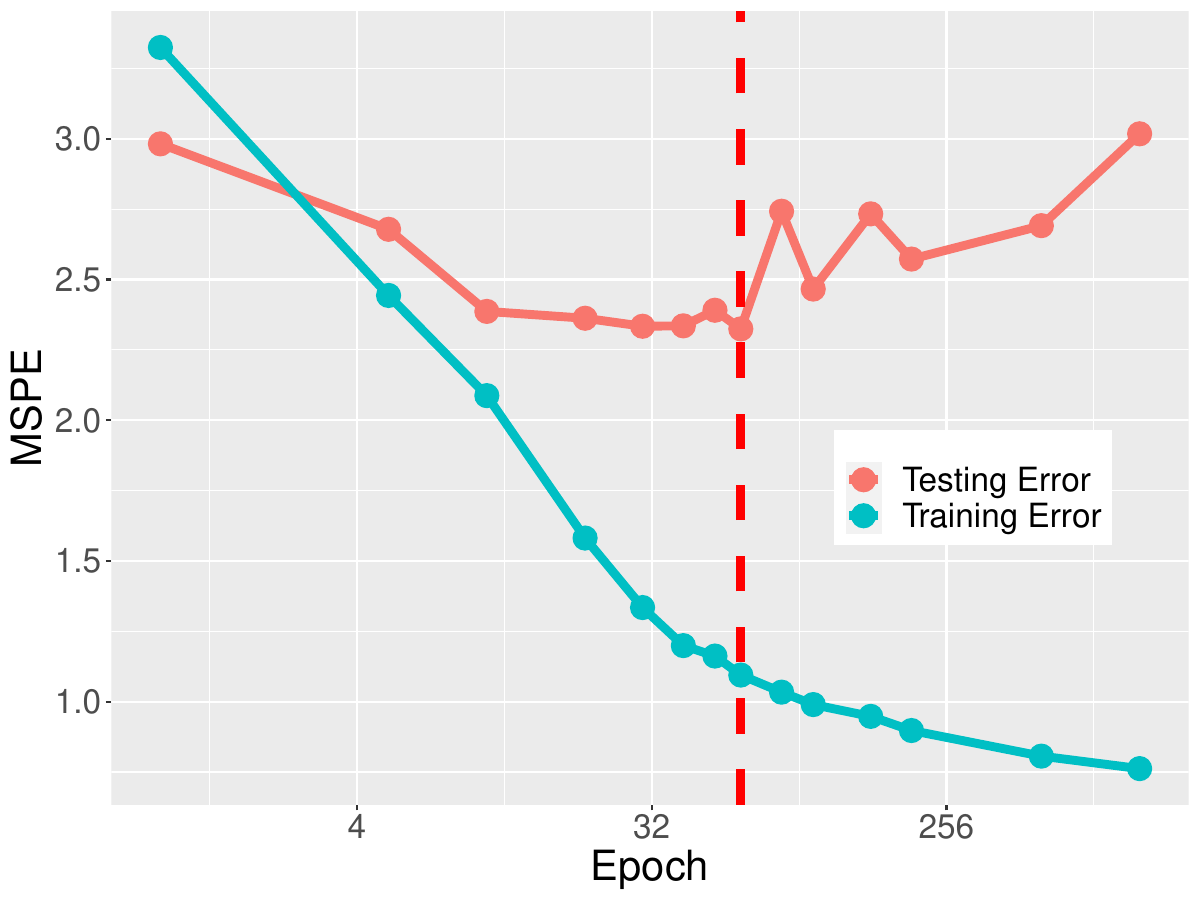}
  \caption{GRU}
  \label{GRU96}
\end{subfigure}
\begin{subfigure}{0.3\textwidth}
  \centering
  \includegraphics[width=1\textwidth,]{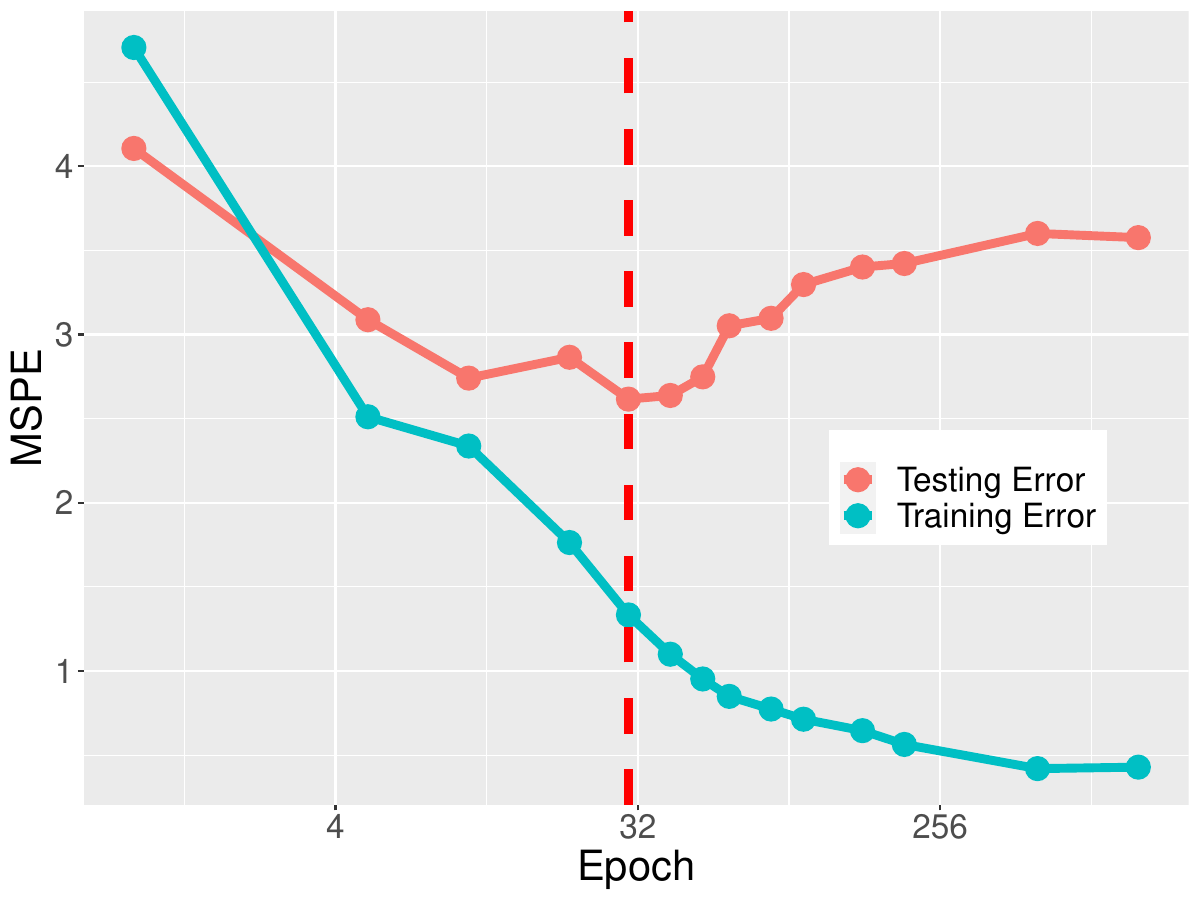}
  \caption{LSTM}
  \label{LSTM96}
\end{subfigure}
\caption{MSPE as a function of the epochs for the (a) GRU and (b) LSTM model on the Lorenz-96 data with $n = 81$ vector elements. The number of epochs in the $x$-axis is expressed in the logarithmic scale. The red dashed vertical line denotes the tuned number of epochs.}
\label{TD96}
\end{figure}

Figure \ref{TD96} shows the diagnostics on the optimal number of epochs for the GRU and LSTM model.

\section*{Section 5: Application}

\begin{figure}[H]
\centering
\begin{subfigure}{0.45\textwidth}
  \centering
  \includegraphics[width=1\textwidth,]{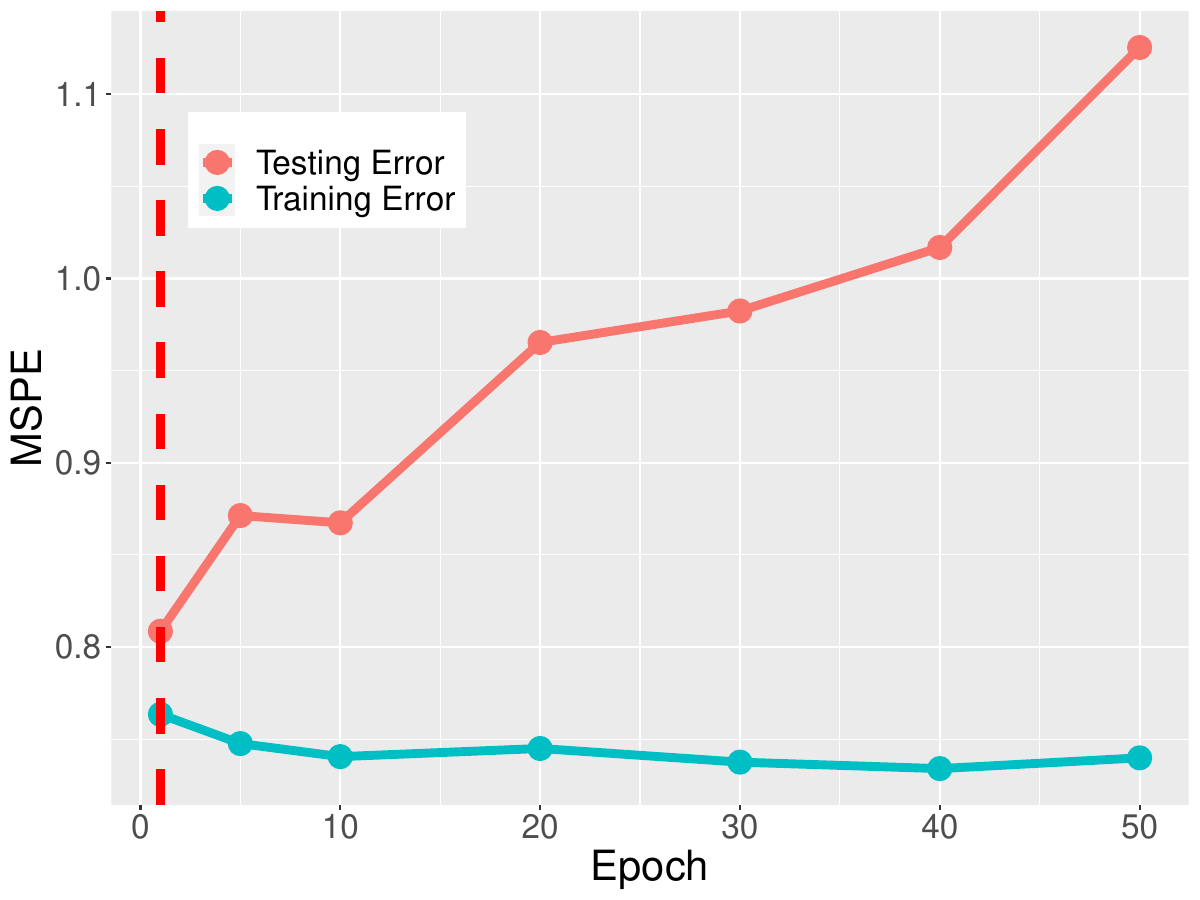}
  \caption{GRU}
  \label{GRU}
\end{subfigure}
\begin{subfigure}{0.45\textwidth}
  \centering
  \includegraphics[width=1\textwidth,]{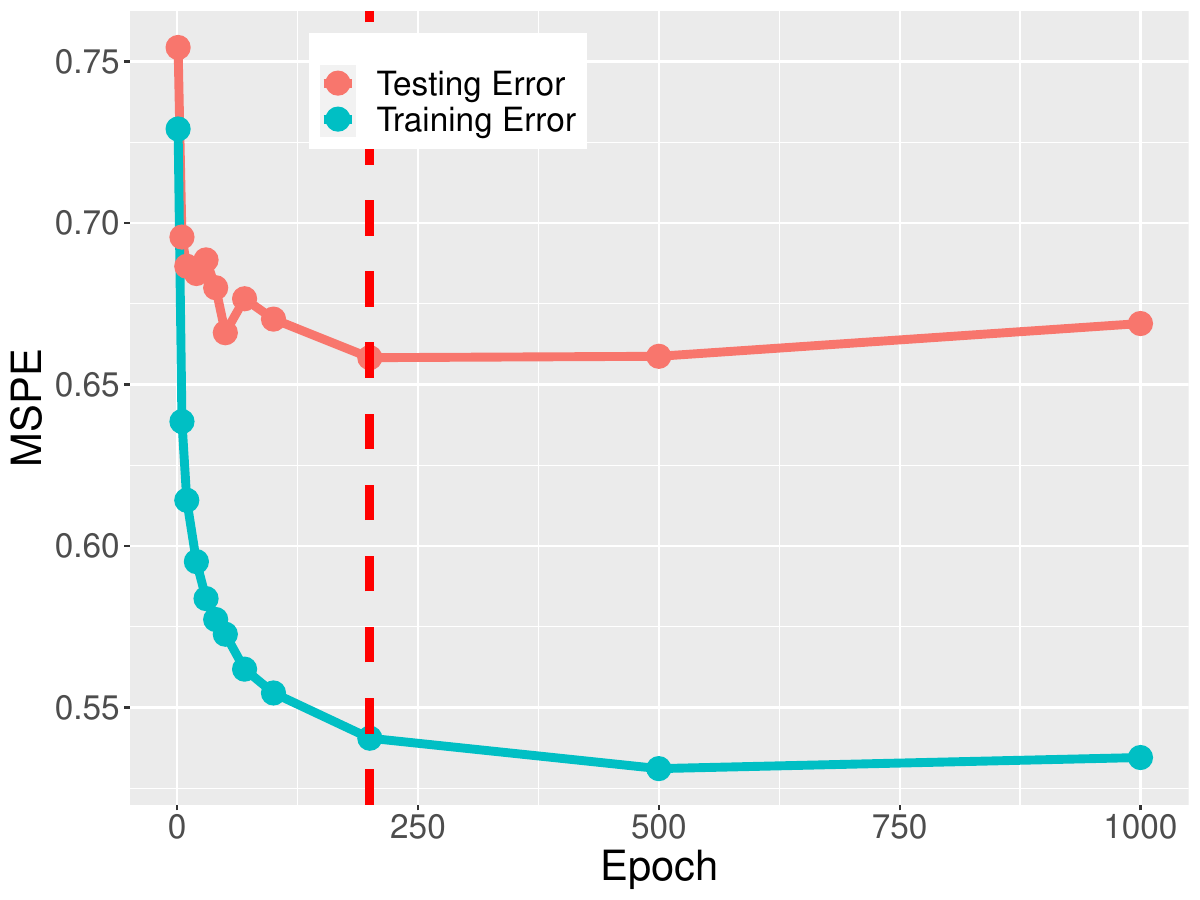}
  \caption{LSTM}
  \label{LSTM}
\end{subfigure}
\caption{MSPE as a function of the epochs for the (a) GRU and (b) LSTM model on the WRF simulated wind speed. The red dashed vertical line denotes the tuned number of epochs.}
\label{TD}
\end{figure}

Figure \ref{TD} shows how the optimal number of epochs is chosen for GRU and LSTM. Table \ref{grid_search} showcases the grid of hyper-parameters for the inference in the deep ESN model. In addition, although it has been shown that for another case study a deep ESN improves prediction accuracy \citep{bon23}, we show in Table \ref{tab:deep_test} that this is not the case for our data. This discrepancy can be at least partially explained by the different aim of our work, as in this application we are interested in much shorter forecasting lead times, where the advantages of a deep models are not as apparent. It can be observed from Table \ref{tab:deep_test} that when the data have 100 locations and 100 time points, ESN with $D=3$ gives the most efficient prediction errors. However, as the dataset increases to 1,000 locations and 1,000 time points, the prediction error increases as $D$ increases.

\begin{table}[H]
    \centering
    \caption{Search grid for $\btheta$ of the ESN.}
    \begin{tabular}{|c|c|c|}
    \hline
       Interpretation  & Notation & Grid  \\
       \hline
       Dimension of the state vector at layer $D$  & $n_{h,D}$ & 1000,1500,\dots,5000 \\ 
       \hline
       Input lag & $m$ & 1,2,\dots,5 \\
       \hline
       Matrix scale parameter at layer $D$ & $\nu_D$ & $0.1,0.2,\dots,1$\\
        \hline
       Ridge penalty  & $\lambda$ & $0.1,0.15,\dots,0.3$\\
         \hline
       Width of the transition weight matrix $\bW_{D}$  & $\eta_{W_D}$ & $0.005, 0.01, 0.05, 0.1, 0.15$\\
                \hline
        Width of the input weight matrix $\bW_{D}^{\text{in}}$  & $\eta_{W_D^{\text{in}}}$ & $0.005, 0.01, 0.05, 0.1, 0.15$\\
                \hline
        Density of the transition weight matrix $\bW_{D}$  & $\pi_{W_D}$ & $0.005, 0.01, 0.05, 0.1, 0.15$\\
                \hline
        Density of the input weight matrix $\bW_{D}^{\text{in}}$  & $\pi_{W_D^{\text{in}}}$ & $0.005, 0.01, 0.05, 0.1, 0.15$\\
                \hline
       Leaking rate  & $\alpha$ & $0.1,0.2,\dots,1$\\
    \hline
    \end{tabular}
    \label{grid_search}
\end{table}

\begin{table}[H]
    \centering
    \caption{Validation errors on the testing data for the ESN model as a function of depth $D$, width $n_h$, as well as the number of randomly sub-sampled locations and time points. }
    \begin{tabular}{|c|c|c|c|c|c|}
    \hline
       Locations  & Time points & $n_h$ & MSE ($D=1$) & MSE ($D=2$) & MSE ($D=3$)  \\
       \hline
        100 & 100 & 50 & 0.592 & 0.598 & 0.523 \\
        \hline
        100 & 100 & 100 & 0.658 & 0.596 & 0.538 \\
        \hline
        500 & 500 & 100 & 0.617 & 0.633 & 0.638 \\
        \hline
        500 & 500 & 200 & 0.664 & 0.682 & 0.642 \\
        \hline
        1000 & 1000 & 200 & 0.564 & 0.644 & 0.661 \\
        \hline
        1000 & 1000 & 400 & 0.567 & 0.659 & 0.801 \\
        \hline
    \end{tabular}
    \label{tab:deep_test}
\end{table}

\begin{table}[H]
    \centering
     \caption{The estimated shrinkage parameter $\hat{\delta}$ according to the best mean coverage in the expanding square defined in Section \ref{sec:app_uq}. The parameter $N$ denotes the number of locations within the square. The numbers marked in red are the shrinkage parameters resulting in the best marginal coverage across all the locations.}
    \begin{tabular}{|c|c|c|c|c|c|c|}
        \hline
        $N$ & 4 & 9 & 49 & 121 & 210 & 323  \\
        \hline
        $\hat{\delta}$ - Lead One & { 0.36} & 0.33 & 0 & 0 & 0 & 0 \\
        \hline
        $\hat{\delta}$ - Lead Two& { 0.54} & 0.62 & 0.15 & 0 & 0 & 0\\
        \hline
        $\hat{\delta}$ - Lead Three &  0.78 & { 0.71} & 0.49 & 0.3 & 0.35 & 0.08\\
        \hline
    \end{tabular}
    \label{shrinkage}
\end{table}

\begin{figure}[H]
    \centering
    \includegraphics[width = 0.8\textwidth,]{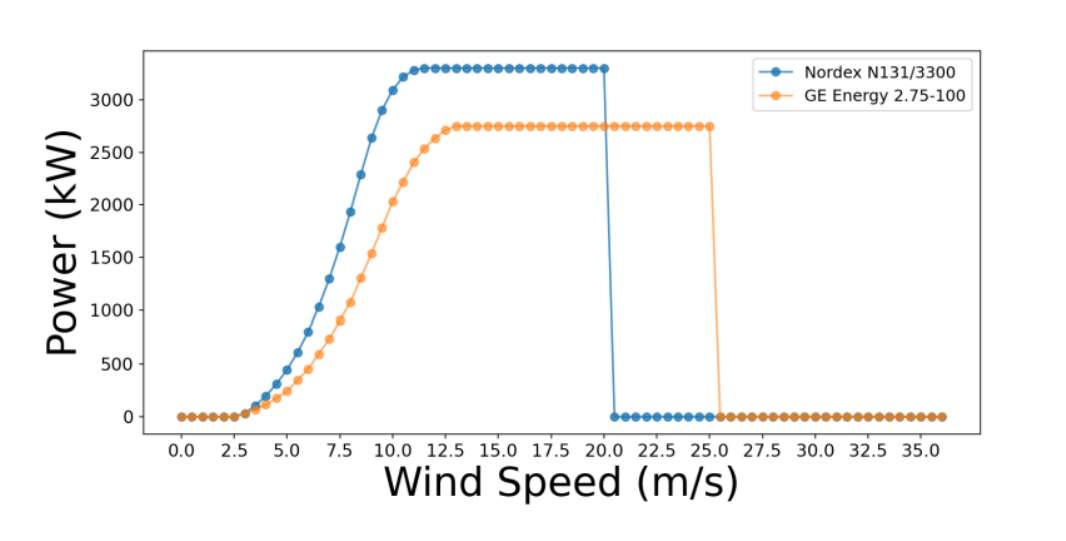}
    \caption{Wind power curve used for transforming wind speed to wind power for two popular turbine models.}
    \label{fig:power_curve}
\end{figure}

\begin{figure}[H]
    \centering
    \includegraphics[width = 0.8\textwidth,]{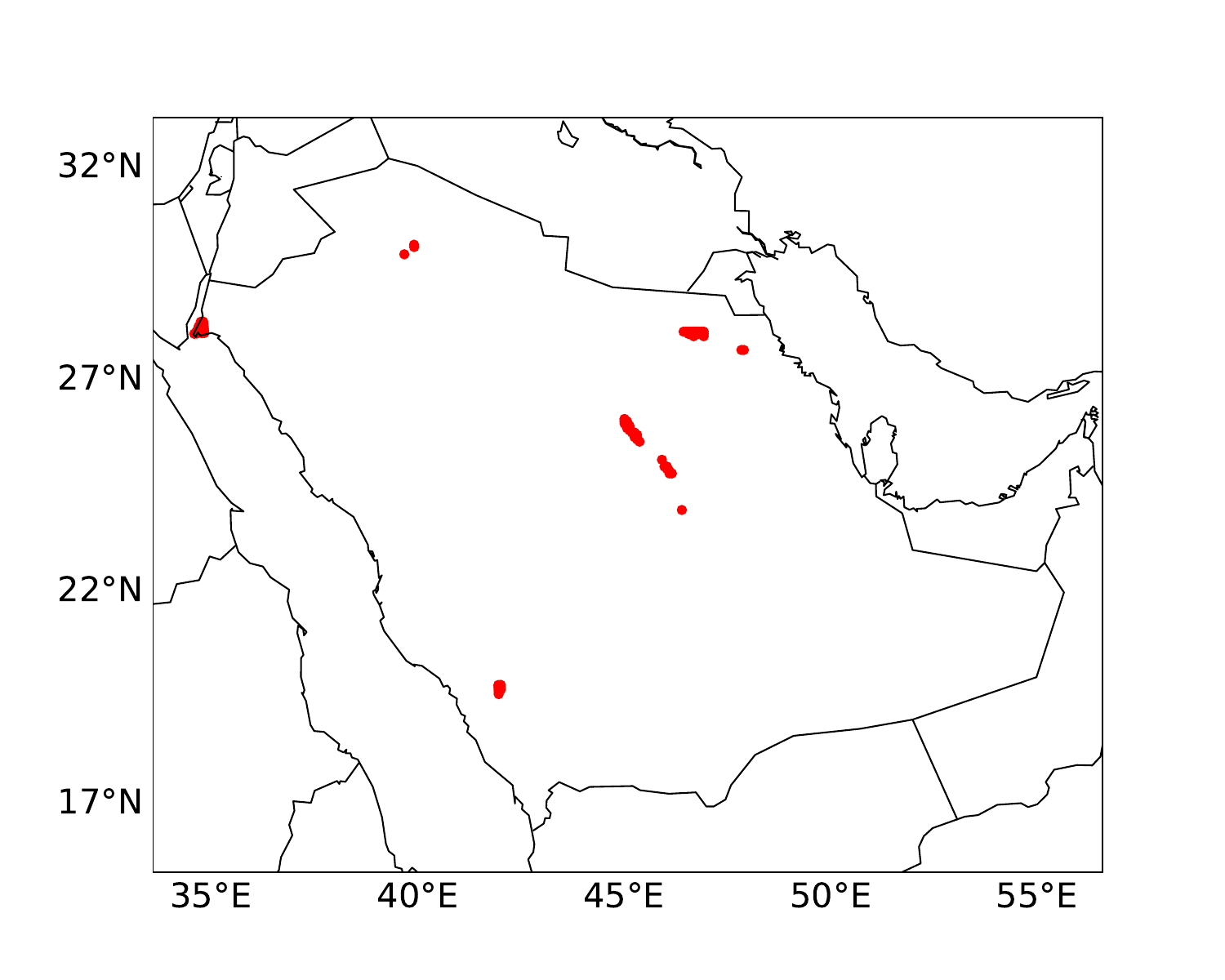}
    \caption{The 75 optimal locations for wind farming in Saudi Arabia, as calculated in \cite{giani2020closing}.}
    \label{fig:power_loc}
\end{figure}

The set of hyper-parameters $\btheta$ that results in the best testing error is summarized in Table~\ref{tab:grid_cv}. Figure \ref{fig:power_curve} shows an example of two power curves for two different turbine models, one from Nordex and one from General Electric (GE). Figure \ref{fig:power_loc} shows the optimal locations for wind farming as indicated in \cite{giani2020closing}.

\begin{table}[H]
    \centering
    \caption{The set of estimated hyper-parameters for the ESN model using cross-validation in the application.}
    \begin{tabular}{|c|c|c|}
    \hline
       Interpretation  & Notation & Grid  \\
       \hline
       Dimension of the state vector at layer $D$  & $n_{h,D}$ & 2500 \\ 
       \hline
       Input lag & $m$ & 1 \\
       \hline
       Matrix scale parameter at layer $D$ & $\nu_D$ & 0.9\\
        \hline
       Ridge penalty  & $\lambda$ & 0.15\\
         \hline
       Width of the transition weight matrix $\bW_{D}$  & $\eta_{W_D}$ & 0.05\\
                \hline
        Width of the input weight matrix $\bW_{D}^{\text{in}}$  & $\eta_{W_D^{\text{in}}}$ & 0.01\\
                \hline
        Density of the transition weight matrix $\bW_{D}$  & $\pi_{W_D}$ & 0.1\\
                \hline
        Density of the input weight matrix $\bW_{D}^{\text{in}}$  & $\pi_{W_D^{\text{in}}}$ & 0.01\\
                \hline
       Leaking rate  & $\alpha$ & 1\\
    \hline
    \end{tabular}
    \label{tab:grid_cv}
\end{table}

\end{document}